\documentclass[journal,10pt,final]{IEEEtran}

\usepackage{graphicx}
\usepackage{epstopdf}
\usepackage{amsthm,amsmath,amssymb}
\usepackage{mathrsfs}
\usepackage{setspace}
\usepackage{amsmath}
\usepackage{booktabs}
\usepackage{enumerate}
\usepackage{cite}
\usepackage{stfloats}
\usepackage{subfigure}

\usepackage{algorithm}
\usepackage{algorithmic}
\usepackage{multirow}
\usepackage{flafter}
\usepackage{bm}

\usepackage{url}
\usepackage{xcolor}
\usepackage{subfigure}
\usepackage{fancyhdr}
\usepackage{mdwmath}
\usepackage{mdwtab}
\usepackage{caption}
\usepackage{amsthm}
\usepackage{setspace}
\usepackage{tabularx}
\usepackage{tabu}
\usepackage{lipsum}
\usepackage{hyperref}

\newtheorem{remark}{Remark}
\allowdisplaybreaks[4]

\ifCLASSINFOpdf
\else
\fi

\makeatletter
\newenvironment{breakablealgorithm}
  {
   \begin{center}
     \refstepcounter{algorithm}
     \hrule height.8pt depth0pt \kern2pt
     \renewcommand{\caption}[2][\relax]{
       {\raggedright\textbf{\ALG@name~\thealgorithm} ##2\par}%
       \ifx\relax##1\relax 
         \addcontentsline{loa}{algorithm}{\protect\numberline{\thealgorithm}##2}%
       \else 
         \addcontentsline{loa}{algorithm}{\protect\numberline{\thealgorithm}##1}%
       \fi
       \kern2pt\hrule\kern2pt
     }
  }{
     \kern2pt\hrule\relax
   \end{center}
  }
\makeatother

\begin{document}

\title{Intelligent Trajectory Design for RIS-NOMA aided Multi-robot Communications}

\author{
{Xinyu~Gao,~\IEEEmembership{Graduate Student Member,~IEEE,}
Xidong~Mu,~\IEEEmembership{Member,~IEEE,}\\
Wenqiang~Yi,~\IEEEmembership{Member,~IEEE,}
and Yuanwei~Liu,~\IEEEmembership{Senior Member,~IEEE}

\thanks{Part of this work has been presented at the IEEE International Conference on Communications, 14-23 June, 2021~\cite{IEEEhowto:XGao}.}
\thanks{X. Gao, X.Mu, W. Yi, and Y. Liu are with the School of Electronic Engineering and Computer Science, Queen Mary University of London, London E1 4NS, U.K. (e-mail:\{x.gao, xidong.mu, w.yi, yuanwei.liu\}@qmul.ac.uk).}
}
}

\maketitle

\begin{abstract}
  A novel reconfigurable intelligent surface-aided multi-robot network is proposed, where multiple mobile robots are served by an access point (AP) through non-orthogonal multiple access (NOMA). The goal is to maximize the sum-rate of whole trajectories for the multi-robot system by jointly optimizing trajectories and NOMA decoding orders of robots, phase-shift coefficients of the RIS, and the power allocation of the AP, subject to predicted initial and final positions of robots and the quality of service (QoS) of each robot. To tackle this problem, an integrated machine learning (ML) scheme is proposed, which combines long short-term memory (LSTM)-autoregressive integrated moving average (ARIMA) model and dueling double deep Q-network (D$^{3}$QN) algorithm. For initial and final position prediction for robots, the LSTM-ARIMA is able to overcome the problem of gradient vanishment of non-stationary and non-linear sequences of data. For jointly determining the phase shift matrix and robots' trajectories, D$^{3}$QN is invoked for solving the problem of action value overestimation. Based on the proposed scheme, each robot holds an optimal trajectory based on the maximum sum-rate of a whole trajectory, which reveals that robots pursue long-term benefits for whole trajectory design. Numerical results demonstrated that: 1) LSTM-ARIMA model provides high accuracy predicting model; 2) The proposed D$^{3}$QN algorithm can achieve fast average convergence; and 3) RIS-NOMA networks have superior network performance compared to RIS-aided orthogonal counterparts. 
  \end{abstract}


\IEEEpeerreviewmaketitle

\vspace{-0.4cm}
\section{Introduction}
\vspace{-0.1cm}
Nowadays, it is commonly convinced that robots are far less capable when operating independently, and the real power lies in the cooperation of multiple robots. Therefore, multi-robot systems in a shared environment have attracted significant attention in terms of various emerging applications, e.g., cargo delivery, automatic patrol, and emergency rescue \cite{Multi-robot Book}. Among these scenarios,  robots are required to coordinate with each other to achieve some well-defined goals, e.g., moving from one given position to another. However, with the increasing complexity of the application environment, the large local computational resources are also consumed when collaborating to handle tasks in the multi-robot system. Due to the cooperation requirement and high computation complexity for the trajectory design in multi-robot systems, wireless communications with advanced multiple access techniques are important for multi-robot systems \cite{IEEEhowto:Batalin}.
\par
As an emerging technique, non-orthogonal multiple access (NOMA) \cite{IEEEhowto:Saito, IEEEhowto:LDai, IEEEhowto:ZDing} adopts a flexible successive interference cancellation (SIC) receiver for robust multiple access. It improves spectrum efficiency by opportunistically exploring users’ channel conditions. However, there still exists a shortage of spectrum in some communication regions, e.g., blind zone. To tackle this problem, reconfigurable reflecting surfaces (RIS) \cite{IEEEhowto:YZhu,IEEEhowto:Wu,Mei1,IEEEhowto:Zhao,IEEEhowto:EBasar,IEEEhowto:JiangboSi} is a potential candidate to improve the spectrum efficiency, which is passive equipment that can proactively reflect the signal to the users. Specifically, the employment of the RIS is able to create a virtual line of sight (LOS) between access points (AP) and robots when robots are located in the communication blind zone. In view of the advantages brought by the RIS and NOMA techniques, RIS-NOMA is regarded as a potential solution to efficiently handle the trajectory design problems of the multi-robot system.

\vspace{-0.35cm}
\subsection{Related Works}

\subsubsection{RIS-NOMA Networks}
RIS-NOMA technique becomes appealing in recent years, and it has been applicated in various scenarios, including spectrum efficiency and user connectivity improvement \cite{IEEEhowto:QChen,IEEEhowto:YLi,IEEEhowto:ZDing1,IEEEhowto:XMU,IEEEhowto:XGAO}, energy consumption decrease \cite{IEEEhowto:ZLi}, etc. In \cite{IEEEhowto:FFang}, alternating successive convex approximation (SCA) and semi-definite relaxation (SDR) based energy-efficient algorithms were proposed to yield a good tradeoff between the sum-rate maximization and total power consumption minimization on RIS-aided NOMA networks, by maximizing the system energy efficiency by jointly optimizing the transmit beamforming at the base station (BS) and the reflecting beamforming at the RIS. To establish stable and high-quality communication links between AP and the robotic users, an indoor robot navigation system was investigated in \cite{IEEEhowto:Xmu}, where RIS was employed to enhance the connectivity and NOMA was adopted to improve the communication efficiency between the AP and robotic users. An SDR-based solution in \cite{IEEEhowto:MZeng} was proposed to address a joint power control at the users and beamforming design at the IRS for maximizing the sum-rate of all users in a RIS-aided uplink NOMA system. Also, the impact of the number of reflecting elements on the sum-rate was revealed. The authors in \cite{IEEEhowto:Ni} proposed a novel framework of resource allocation in multi-cell RIS-aided NOMA networks, which was capable of being enhanced with the aid of the RIS, and the proper location of the RIS is also guarantee the trade-off between spectrum and energy efficiency. In order to examine the effectiveness of RIS in the NOMA system with respect to transmitting power consumption, the authors in \cite{IEEEhowto:HWANG} explored the relationship between individual user's transmit power and the variables of phase shifts, and solved the phase shift determination problem via the sequential rotation algorithm.

\subsubsection{Communication-connected Multi-robot Trajectory Design}
Leveraging the wireless communication technology is capable of providing services for the communication-connected multi-robot systems that require communication quality, so as to provide optimal trajectory design in special environments \cite{IEEEhowto:YLiu5}. To avoid path conflicts between robots and repeated exploration and information collected from the same region by different robots, a novel algorithm was proposed \cite{10}. Among them, decisions to select locations for exploration and information collection were guided by a utility function that combines Gaussian Process-based distributions for information entropy and communication signal strength, along with a distributed coordination protocol. The authors of \cite{11} formulated the problem of multi-robot informative path planning under continuous connectivity constraints as an integer program leveraging the ideas of bipartite graph matching and minimal node separators. In \cite{12}, a framework for planning and perception for multi-robot exploration in large and unstructured three-dimensional environments was presented. A Gaussian mixture model for global mapping was employed in the proposed method to model complex environment geometries while maintaining a small memory footprint which enables distributed operation with a low volume of communication. In \cite{13}, the authors developed a communication-based navigation framework for multi-robot systems in unknown areas that solely exploit the sensing information and shared data among the agents. Additionally, the millimeter-wave \cite{14} and multi-input multi-output (MIMO) \cite{15} technologies were integrated in the multi-robot trajectory design. In contrast to conventional multi-robot path planning, the authors of \cite{16} defined a type of multi-robot association-path planning problem aiming to jointly optimize the robots' paths and the robots-AP associations. Using geometrically motivated assumptions and characteristics of MIMO, the authors in \cite{17} derived transmitter spacing rules that can be easily added to existing path plans to improve backhaul throughput for data offloading from the robot team, with minimal impact on other system objectives.

\vspace{-0.2cm}
\subsection{Motivations and Contributions}
Although mentioned studies have revealed the potential abilities of communication technology in multi-robot systems, the research on the communication-connected multi-robot long-term trajectory design is in its early stage. For instance, the previous research mainly focuses on the trajectory design constrained by the geographic environment, ignoring the important role of communication quality in multi-robot trajectory design. Since the ultimate goal for the multi-robot system is to achieve specific goals such as simultaneous localization and mapping or moving to some target areas with minimum time/energy consumption \cite{18,19}, the achievable maximum performance (i.e., capacity) of the whole system is a basis for the multi-robot system to efficiently achieve these goals. Additionally, the exploration of the long-term benefits of the dynamic movement of mobile robots in networks is also neglected in previous research contributions. The limitations and challenges are summarized as follows:
\begin{itemize}
  \item \textbf{The characterization of the channel model}: The signal may be blocked in some areas by the obstacles, while the channel model also changes abruptly. As result, the position-dependent channel model indicates that the channel model characterization is challenging. 
  \item \textbf{The determination of initial and final positions}: In some special scenarios (e.g., fixed-point cruise), different initial and final positions bring various trajectory designs and obtain miscellaneous achievable sum-rate. The trajectory with a high sum-rate is able to make the robot exchange much environment information with the controller and further effectively promote subsequent operation. Therefore, it's important to find the optimal initial and final positions corresponding to the trajectories with maximum sum-rate.
  \item \textbf{The influence of the resource allocation strategy}: To realize optimal trajectories, the robots need to continuously evaluate the previous network performance and resource allocation strategy at each timestep to reap the rewards and determine the next action. Therefore, the robots continuously interacting with the environment is also challenging.
  \item \textbf{The design of the ML-based algorithm}: Since the positions of robots are determined by the previous positions and environments, conventional algorithms fail to handle a time-varying-based MDP, and vanilla ML algorithms have limitations to achieve good performance. Thus, compared to these algorithms, the vanilla ML algorithms need to be further improved for obtaining better performance than the vanilla algorithms.
\end{itemize}
\par
In response to the above limitations and challenges, we proposed a novel ML framework, which integrated Long short-term memory (LSTM)-autoregressive integrated moving average (ARIMA) and dueling double deep Q-network (D$^{3}$QN) algorithms to give full play to the conventional ML algorithms. LSTM-ARIMA algorithm is able to efficiently handle non-stationary \cite{non-stationary} and non-linear sequences of any size of data and completely solve the problem of gradient vanishment \footnote{For LSTM, the memory neural networks are introduced to train a model, where the non-stationary data with dramatic fluctuation can be handled. For ARIMA, it is to utilize the difference operation to convert the non-stationary sequences to stationary sequences first and then handle the stationary sequences based on the AR model.}. D$^{3}$QN algorithm has the capability of solving the problem of action value overestimation in double DQN and dueling DQN algorithms and improving the accuracy of action selection for each state. After providing the range of the initial and final positions by the LSTM-ARIMA algorithm, the D$^{3}$QN algorithm is able to characterize the channel models and further provide the optimal initial and final positions, trajectories, and phase shift design for the robots. The main contributions of this paper are summarized as follows:
\begin{itemize}
  \item We propose a new framework for RIS-NOMA-aided multi-robot networks. The RIS is employed to enhance communication efficiency, by proactively reflecting the incident signals, while NOMA is invoked for improving the spectrum efficiency of the multi-robot system. Based on the framework, an optimization problem is formulated to obtain the maximum sum-rate of the whole trajectories for all robots. by jointly optimizing trajectories for robots, reflecting coefficient matrix of RIS, successive interference cancelation (SIC) decoding order for NOMA, power allocation at the AP, subject to the quality of service (QoS) for the robots.
  \item We adopt the LSTM-ARIMA algorithm for training a more accurate model to guide the optimal trajectory design for the robots. LSTM-ARIMA algorithm is able to handle non-stationary and non-linear sequences of any size of data, as well as completely solve the problem of gradient vanishment. For channels, at each point/time, the experimental observations of small-scale fading of channels are different, meaning they have different sequence properties (e.g., mean and variance). Hence, the observations at different points/times are non-stationary. The LSTM-ARIMA model is able to unify small-scale fading properties by difference operation to further mimic the small-scale finding distribution in the considered environment.
  \item We demonstrate that the proposed D$^{3}$QN algorithm is capable of providing the optimized robots' trajectories, phase shifts of RIS, and optimal initial and final positions for the robots, according to the obtained results of the LSTM-ARIMA algorithm. The D$^3$QN algorithm is an online reinforcement learning (RL) algorithm to train an off-policy, which fully reaps the merits of double DQN and dueling DQN to introduce the target network and split the network structure to obtain the accurate Q value. After numerous repetitive training, the performance of the D$^{3}$QN algorithm is proved to outperform double DQN and dueling DQN algorithms.
  \item We demonstrate that the proposed D$^{3}$QN algorithm efficiently solves the trajectory design problem of the multi-robot system. Specifically, the D$^{3}$QN algorithm is able to achieve the fastest convergence speed than conventional double DQN and dueling DQN algorithms. Furthermore, when applying the D$^{3}$QN, dueling DQN and double DQN algorithms on trajectory design upon different elements of RIS, the proposed D$^3$QN algorithm is able to find a shorter total path than the conventional dueling DQN and double DQN algorithms. Compared to the RIS-OMA technique, the RIS-NOMA technique is able to achieve a higher communication sum-rate with a shorter traveling distance for robots, which verifies the effectiveness of the NOMA technique. The optimal decoding order has better performance than the fixed decoding order and random decoding order, which also shows that the decoding order is a considerable factor in NOMA networks.
\end{itemize}

\vspace{-0.4cm}
\subsection{Orgainizations}
The rest of this paper is organized as follows. Section II presents the system model for the considered RIS-aided multi-robot NOMA networks, and the passive beamforming and trajectory design problems are formulated. In Section III, we propose the LSTM-ARIMA model and D$^{3}$QN algorithm, which is employed to predict the initial position and the final position, jointly planning trajectories and designing the beamforming. Section IV presents numerical results to verify the effectiveness of the proposed machine learning-based optimization algorithms for joint trajectories planning and passive beamforming design, as well as the performance of the algorithms. Finally, Section V concludes this paper.
\par
\emph{Notations:} Scalars, vectors, and matrices are denoted by lower-case, bold-face lower-case, and bold-face upper-case letters, respectively. $\mathbb{C}^{K \times N}$ denotes the space of $K \times N$ complex-valued vectors. The conjugate transpose of vector $\mathbf{a}$ is denoted by $\mathbf{a}^{H}$. diag($\mathbf{a}$) denotes a diagonal matrix with the elements of vector $\mathbf{a}$ on the main diagonal. $|\mathbf{a}|$ denotes the norm of vector $\mathbf{a}$. $*$ denotes the dot multiplication operation. log$_{2}$($\mathbf{A}$) represent a logarithmic function with a constant base of 2 for matrix $\mathbf{A}$.

\vspace{-0.4cm}

\section{System Model and Problem Formulation}

\subsection{System Model}
\begin{figure}[htbp] 
  \centering
  \setlength{\belowcaptionskip}{-0.3cm}
  \includegraphics[scale = 0.12]{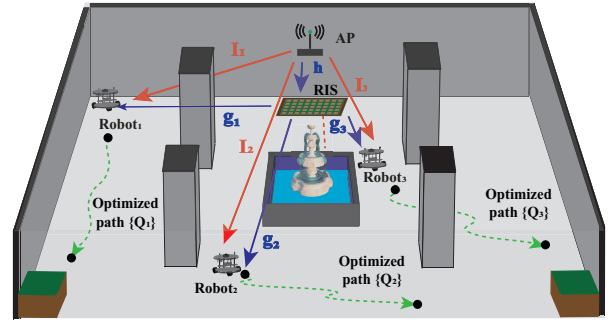}
  \caption{Illustration of the RIS aided multi-robots cruise system for the indoor environment.}
  \label{Illustration of the RIS aided multi robots cruise system for the indoor environment}
\end{figure}
As shown in Fig.~\ref{Illustration of the RIS aided multi robots cruise system for the indoor environment}, we focus our attention on a downlink RIS-aided multi-robot NOMA network, where one single-antenna AP serves \emph{$\mathcal{X}$} single-antenna mobile wheeled robots, assisting by a RIS with \emph{K} passive reflecting elements. The passive reflecting elements in the RIS can be partitioned into \emph{M} sub-surfaces, while each sub-surfaces consists of \emph{$\widetilde{K}=K/M$} elements. Assume that the two-dimensional (2D) horizontal ground space for robots moving is approximately smooth without undulation, indicating that the height \emph{$h_{r}$} of the robot (the height of the antenna) is regarded as a constant value. The other values' (e.g., velocity, turning radius) adjustment involved in the robot's mechanical motion are also considered to be accurate. Additionally, the receiver is located on the top of the robots. Define a three-dimensional (3D) Cartesian coordinate system, the positions of AP and robot \emph{$i$} are denoted as (\emph{$x_{A}$},\emph{$y_{A}$},\emph{$h_{A}$}) and (\emph{$x_{i}$},\emph{$y_{i}$},\emph{$h_{r}$}), respectively. Note that for users' fairness, the RIS is arranged in the center of the ceiling in the motion space, where the position can be denoted as (\emph{$x_{I}$},\emph{$y_{I}$},\emph{$h_{I}$}).
\par
In view of the deployment of the RIS, the composite received signal of each robot is the combination of two components, which are derived from the AP-robot direct link, and the AP-RIS-robot reflecting link. Denote the baseband equivalent channels from the AP to \emph{$i$}-th robot, the AP to RIS and the RIS to \emph{$i$}-th robot as \emph{$\mathbf{l}_{i}$}, \emph{$\mathbf{h}^{H} \in \mathbb{C}^{1 \times M}$}, and \emph{$\mathbf{g}_{i} \in \mathbb{C}^{M \times 1}, i=\{1,2,\cdots,\mathcal{X}\}$}, respectively. For the distance-reference path loss, it can be modeled as \emph{$L_{u} = Cd_{u}^{-\gamma}, u=\{Ai,Ri,AI\}$} \cite{IEEEhowto:QWu}, where \emph{C} denotes the path loss at the reference distance of 1 meter, \emph{$d_{[\cdot]}$} is distance for individual link, \emph{$\gamma$} represents the path loss factor, respectively. Since the positions of robots are time-sensitive, the positions of \emph{i}-th robot can be re-denoted as \emph{$q_{i}(t) = (x_{i}(t), y_{i}(t), h_{r})$}. For small-scale fading, we assume a Rician fading channel models for the BS to \emph{i}-th robot, the RIS to \emph{i}-th robot channels and the AP to RIS channel. Thus, the three individual channels for AP to \emph{i-th} robot, RIS to \emph{i-th} robot and AP to RIS can be expressed as
\par
\vspace{-0.2cm}
\noindent
\begin{align}\label{1}
  \mathbf{l}_{i}(q_{i}(t)) = &L_{Ai}(q_{i}(t)) \{\sqrt{\frac{1}{\alpha_{Ai}(q_{i}(t))+1}} \nonumber \\
  &\hspace{2.9em} \cdot[\sqrt{\alpha_{Ai}(q_{i}(t))}\tilde{\mathbf{l}}_{i}^{{\rm LOS}}(q_{i}(t))+\hat{\mathbf{l}}_{i}^{{\rm NLOS}}]\},
\end{align}
\par
\vspace{-0.3cm}
\noindent
\begin{align}\label{2}
  \mathbf{g}_{i}(q_{i}(t)) = &L_{Ri}(q_{i}(t)) \{\sqrt{\frac{1}{\alpha_{Ri}(q_{i}(t))+1}} \nonumber \\
  &\hspace{1em} \cdot[\sqrt{\alpha_{Ri}(q_{i}(t))}\tilde{\mathbf{g}}_{i}^{{\rm LOS}}(q_{i}(t))+\hat{\mathbf{g}}_{i}^{{\rm NLOS}}]\},
\end{align}
\begin{align}\label{3}
  \mathbf{h} = L_{AI} \{\sqrt{\frac{1}{\alpha_{AI}+1}}[\sqrt{\alpha_{AI}}\tilde{\mathbf{h}}^{{\rm LOS}}+\hat{\mathbf{h}}^{{\rm NLOS}}]\},
\end{align}
\par
\vspace{-0.2cm}
\noindent
where \emph{$\alpha_{u}, u=\{Ai,Ri,AI\}$}, \emph{$z^{{\rm LOS}}, z=\{\tilde{\mathbf{l}}_{i},\tilde{\mathbf{g}}_{i},\tilde{\mathbf{h}}\}$}, \emph{$w^{{\rm NLOS}}, w=\{\hat{\mathbf{l}}_{i},\hat{\mathbf{g}}_{i},\hat{\mathbf{h}}\}$} denote the Rician factor, deterministic LOS component and random non-line-of-sight (NLOS) Rayleigh fading components, respectively. Here, we are interested in the expected values to obtain the NLOS component. When $\alpha_{Ai}(q_{i}(t)) = 0$ or $\alpha_{Ri}(q_{i}(t)) = 0$, the AP to \emph{i}-th robot link or the RIS to \emph{i}-th robot link is blocked. Otherwise, $\alpha_{Ai}(q_{i}(t)) = \overline{a}$ or $\alpha_{Ri}(q_{i}(t)) = \overline{a}$, where $\overline{a}$ is a constant value. With respect to RIS, denote \emph{$\mathbf{\Phi}(t) = {\rm diag}(\phi_{1}(t),\phi_{2}(t),\cdots,\phi_{M}(t))$} as the reflection coefficients matrix of the RIS, where $\phi_{m}(t)$ = $\beta_{m}(t)e^{j\theta_{m}(t)}$, $k=\{1,2,3,\cdots, M\}$ denotes the reflection coefficient of \emph{m}-th sub-surface of the RIS. Among them, the \emph{$|\beta_{m}(t)| = 1$} and \emph{$\theta_{m}(t) \in [0,2\pi)$} denote the amplitude and phase of \emph{m}-th sub-surface in the RIS. Thus, the effective channel from the AP to the robot \emph{$i$} is given by
\par
\vspace{-0.2cm}
\noindent
\begin{align}\label{4}
  \mathbf{H}_{i}(q_{i}(t)) = \mathbf{h}^{H} \mathbf{\Phi}(t) \mathbf{g}_{i}(q_{i}(t)) + \mathbf{l}_{i}(q_{i}(t)), i=\{1,2,\cdots,\mathcal{X}\}.
\end{align}
\par
\vspace{-0.2cm}
According to the fairness principle, the interference among robots should be considered while one AP serves $\mathcal{X}$ robots simultaneously. We consider the NOMA technology to mitigate interference among robots and for sharing the same time/frequency resources to all the robots. In the NOMA technology, the superposition coding (SC) method is applied at the AP. Let \emph{$S_{i}=\sqrt{p_{i}}s_{i}$} denote the transmitted signal for the robot \emph{$i,i=\{1,2,\cdots,\mathcal{X}\}$}. \emph{$s_{i}$} represents the transmitted information symbol for the robot \emph{$i$}. It is worth noting that \emph{$S_{i}$} is satisfied \emph{$\mathbb{E}[|S_{i}|^{2}] = p_{i} \leq P_{i},i=\{1,2,\cdots,\mathcal{X}\}$}, with \emph{$p_{i}$} and \emph{$P_{i}$} denoting the transmitted power and its maximum value of the robot \emph{$i$}, respectively. The successive interference cancelation (SIC) method is applied for each robot to remove the interference. The robots with stronger channel power gain decode signals of other robots with weaker channel power gain priority over decoding their own signal. The larger power will be allocated to the weak user first while the smaller power will allocate to the strong user. Denote \emph{$O(i)$} as the decoding order of the robot \emph{$i$}. For any two robots \emph{$i$} and \emph{$j$}, \emph{$i \neq j, \hspace{0.25em} i,j=\{1,2,\cdots,\mathcal{X}\}$}, if the decoding order satisfying \emph{$O(i) < O(j)$}, the recieved signal of robot \emph{$i$} can be modeled as
\par
\vspace{-0.3cm}
\noindent
\begin{align}\label{5}
  Y_{i}(q_{i}(t)) = \mathbf{H}_{i}(q_{i}(t))S_{i}(t) + \sum_{O(j) > O(i)} \mathbf{H}_{i}(q_{i}(t))S_{j}(t) + n,
\end{align}
\par
\vspace{-0.3cm}
\noindent
where the \emph{$n \sim \mathcal{CN}(0,\sigma^{2})$} denotes the additive white Gaussian noise (AWGN) with average power \emph{$\sigma^{2}$}. For each robot \emph{$i,i=\{1,2,\cdots,\mathcal{X}\}$}, the achievable rate can be denoted as \emph{$R_{i}$}. To simplify the problem, denote the equivalent channel of AP-RIS-robot link  $\mathbf{h}^{H} \mathbf{\Phi}(t) \mathbf{g}_{i}(q_{i}(t)) = (\mathbf{\upsilon}(t))^{H}\mathbf{\psi}(q_{i}(t))$, where $\mathbf{\psi}(q_{i}(t))$ = diag\{\emph{$(\mathbf{h})^{H}$\}$\mathbf{g}_{i}(q_{i}(t))$}, \emph{$\mathbf{\upsilon}(t) = [\upsilon_{1}(t), \upsilon_{2}(t), \cdots, \upsilon_{M}(t)]^{H}$}, and \emph{$\upsilon_{m} = e^{j\theta_{m}}$}. so the signal-to-interference-plus-noise ratio (SINR) of robot \emph{$i$} is given by
\par
\vspace{-0.3cm}
\noindent
\begin{align}\label{6}
  \tau_{i}(q_{i}(t)) = \frac{|(\mathbf{\upsilon}(t))^{H}\mathbf{\psi}(q_{i}(t))+\mathbf{l}_{i}(q_{i}(t))|^{2}p_{i}(t)}{\sum\limits_{O(j) > O(i)}|(\mathbf{\upsilon}(t))^{H}\mathbf{\psi}(q_{i}(t))+\mathbf{l}_{i}(q_{i}(t))|^{2}p_{j}(t)+\sigma^{2}},
\end{align}
\par
\vspace{-0.3cm}
\noindent
where the \emph{$\sigma^{2}$} denotes the variance of the AWGN. Then, according to $\mathrm{R}=\mathrm{log}_{2}(1+\mathrm{SINR})$, the achievable communication rate at robot \emph{$i$} can be expressed as
\par
\vspace{-0.3cm}
\noindent
\begin{align}\label{7}
  &R_{i}(q_{i}(t)) = {\rm log}_{2}(1+ \nonumber \\
  &\hspace{1.5em}\frac{|(\mathbf{\upsilon}(t))^{H}\mathbf{\psi}(q_{i}(t))+\mathbf{l}_{i}(q_{i}(t))|^{2}p_{i}(t)}{\sum\limits_{O(j) > O(i)}|(\mathbf{\upsilon}(t))^{H}\mathbf{\psi}(q_{i}(t))+\mathbf{l}_{i}(q_{i}(t))|^{2}p_{j}(t)+\sigma^{2}}).
\end{align}
\par
\vspace{-0.3cm}
We consider perfect SIC in our system model, to guarantee rate fairness between two robots, the conditions $R_{i \rightarrow j} \geq  R_{i \rightarrow i}$ should be satisfied under given decoding order $O(i) < O(j)$. The $R_{i \rightarrow j}$ and $R_{i \rightarrow i}$ can be expressed as follows:
\par
\vspace{-0.3cm}
\noindent
\begin{align}\label{2ritoj}
  &R_{i \rightarrow j}(q_{i}(t)) = {\rm log}_{2}(1+ \nonumber \\
  &\hspace{1.5em}\frac{|(\mathbf{\upsilon}(t))^{H}\mathbf{\psi}(q_{j}(t))+\mathbf{l}_{j}(q_{j}(t))|^{2}p_{i}(t)}{\sum\limits_{O(k) > O(i)}|(\mathbf{\upsilon}(t))^{H}\mathbf{\psi}(q_{j}(t))+\mathbf{l}_{j}(q_{j}(t))|^{2}p_{k}(t)+\sigma^{2}}),
\end{align}
\par
\vspace{-0.3cm}
\noindent
\begin{align}\label{2ritoi}
  &R_{i \rightarrow i}(q_{i}(t)) = {\rm log}_{2}(1+ \nonumber \\
  &\hspace{1.5em}\frac{|(\mathbf{\upsilon}(t))^{H}\mathbf{\psi}(q_{i}(t))+\mathbf{l}_{i}(q_{i}(t))|^{2}p_{i}(t)}{\sum\limits_{O(k) > O(i)}|(\mathbf{\upsilon}(t))^{H}\mathbf{\psi}(q_{i}(t))+\mathbf{l}_{i}(q_{i}(t))|^{2}p_{k}(t)+\sigma^{2}}).
\end{align}
\par
\vspace{-0.3cm}

\vspace{-0.5cm}
\subsection{Problem Formulation}
Our goal is to maximize the sum-rate of all robot trajectories by jointly optimizing trajectories for robots, reflection coefficient matrix of RIS, SIC decoding order for NOMA, power allocation at the AP, subject to the QoS for each robot. Hence, the optimization problem is formulated as
\par
\vspace{-0.2cm}
\noindent
\begin{align}
  \max_{\mathbf{\upsilon},\Omega,\{p_{i}\},\varepsilon_{i}, \xi_{i}, \mathbf{Q}} \hspace*{1em}& \sum\limits_{i=1}^{\mathcal{X}} R_{i}(\mathbf{Q}_{i}) \label{8}\\
  {\rm s.t.} \hspace*{1em}& R_{i}(q_{i}(t)) \geq \overline{R}, \hspace*{0.5em} \forall t \in [0, T], \tag{\ref{8}{a}} \label{8a}\\
  & |\upsilon_{m}(t)| = 1, \hspace*{0.5em} \forall m \in \{1, 2, \cdots, M\}, \nonumber \\
  &\hspace{5.75em} \forall t \in [0, T], \tag{\ref{8}{b}} \label{8b}\\
  & \Omega \in \mathbf{\Pi}, \tag{\ref{8}{c}} \label{8c}\\
  & \sum_{i=1}^{\mathcal{X}}p_{i}(t) \leq \mathcal{P}, \tag{\ref{8}{d}} \label{8d} \\
  & q_{i}(0) = \varepsilon_{i}, q_{i}(T) = \xi_{i} \tag{\ref{8}{e}} \label{8e} \\
  & |\dot{q}_{i}(t)| = V, \hspace*{0.5em} \forall t \in [0, T], \tag{\ref{8}{f}} \label{8f} \\
  & q_{i}(t) \in \mathbf{Q}_{i}, \hspace*{0.5em} \forall t \in [0, T], \tag{\ref{8}{g}} \label{8g}
\end{align}
\par
\vspace{-0.2cm}
\noindent
where the \emph{$\overline{R}$}, \emph{$\varepsilon_{i}$}, \emph{$\xi_{i}$}, \emph{$\mathbf{\Pi}$}, and \emph{$\mathbf{Q} = [\mathbf{Q}_{1},\mathbf{Q}_{2},\cdots,\mathbf{Q}_{\mathcal{X}}]$} denote the minimal required communication rate of all the robots, the possible initial positions for the robots, the possible final positions for the robots, the set of all the possible decoding orders, and the set of trajectories for all the robots, respectively. Constraint (\ref{8a}) and constraint (\ref{8b}) are the QoS requirements for robot \emph{$i$} and the restraint for the RIS reflection coefficients. Constraint (\ref{8c}) is the decoding conditions for the NOMA scheme. Constraint (\ref{8d}) is the total transmission power constraint. Constraint (\ref{8e}) and constraint (\ref{8f}) characterize the initial position and final position for each robot and constant velocity for all robots. However, the main difficulty in solving the problem (10) can be summarized in the following reasons. Firstly, the individual instantaneous channel is modeled according to the position of the robot, while it is highly coupled with transmit power allocation and RIS phase shift. Hence, these variables have to be optimized simultaneously instead of decoupling these variables and optimizing them successively. Secondly, the optimal robot trajectories need to be explored by choosing different initial and final positions, while the trajectory for each robot is not always the shortest path from \emph{$\varepsilon_{i}$} to \emph{$\xi_{i}$}, which constrained by the simultaneous arriving conditions for all the robots. Thirdly, the current positions of robots are determined by the previous positions and the environment, which follows the MDP. Fourthly, the experimental observations of small-scale fading of channels are different, the properties of small-scale fading should be stable to guide the trajectories design. The conventional optimization algorithms may fail to be promoted to MDP scenarios. However, our goal is to provide a policy for the whole system, which pursues a long-term effect. Therefore, conventional optimization methods are not proper to be employed to solve these difficulties. The machine learning methods can be invoked to optimize studied RIS-assisted NOMA networks and obtain the optimized trajectories. Additionally, for the implementation of optimal SIC decoding order, the strong users decode the weak users' signals, while the weak users decode the received signal directly without SIC. Therefore, the optimization problem with SIC decoding order constraint becomes a combinatorial optimization problem, where the optimal solution is difficult to obtain by employing the ML algorithm in the current stage \cite{ZDing}. Therefore, in this paper, we apply the exhaustive search method \cite{junzhang} to explore the optimal SIC decoding order.

\vspace{-0.4cm}
\section{Proposed Solutions}
\begin{figure}[htbp]
  \setlength{\belowcaptionskip}{-0.3cm}  
  \centering
  \includegraphics[scale = 0.14]{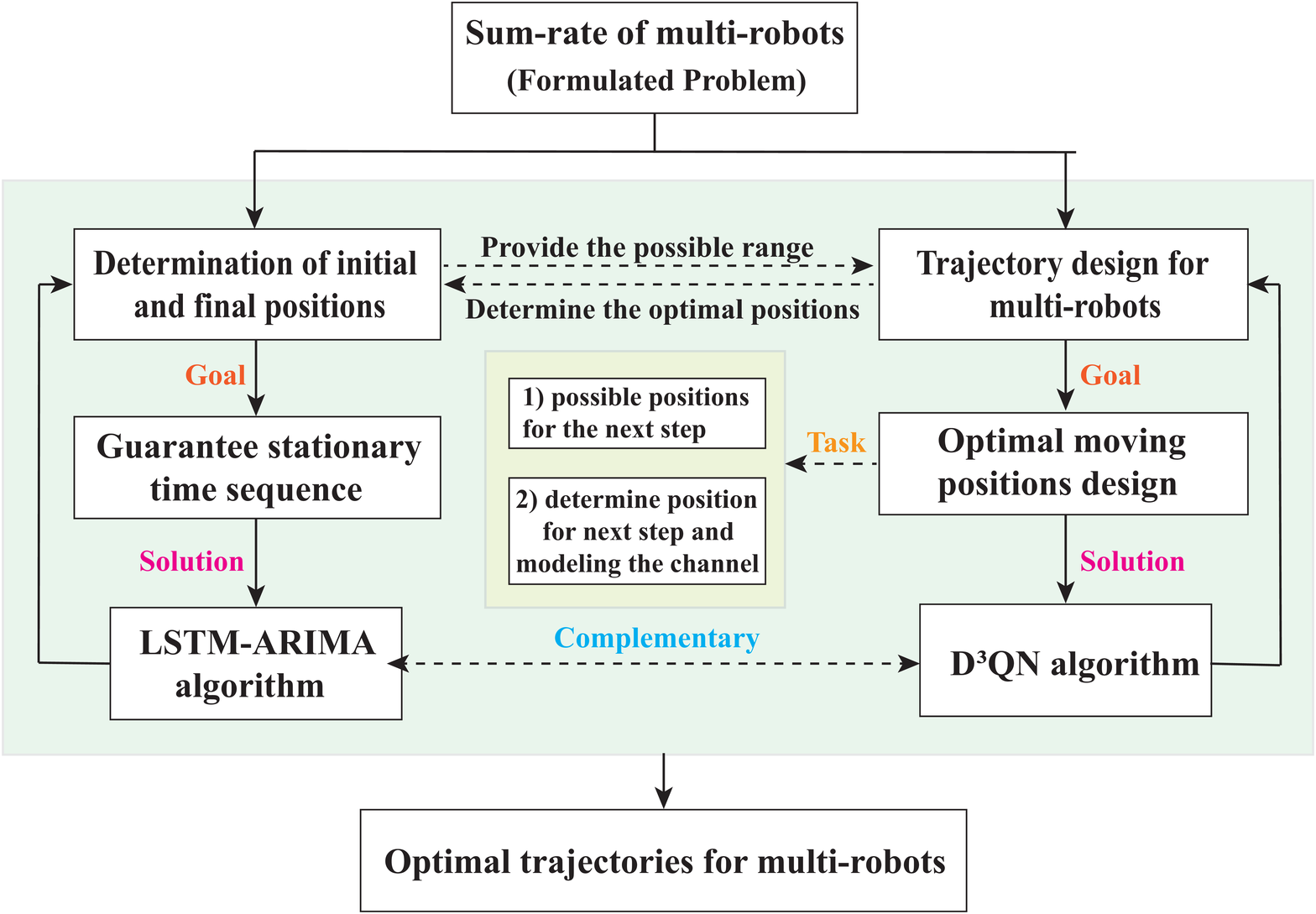}
  \caption{Schematic for solving formulated problem.}
  \label{scheme}
\end{figure}

In this section, we propose an ML-based scheme to solve the problem \eqref{8}, as shown in Fig.~\ref{scheme}. To characterize the non-stationary property in practical robotic communications, we model the sequence property of small-scale fading at different times/locations with the same Rician fading but different parameters in this work as a case study. It is worth mentioning that other non-stationary models can be solved via the same method. Regarding channel statistic change, we assume it changes every coherence time and the decision will be made within each coherence time. Additionally, assuming that the initial position and the final position have a one-to-one correspondence, we also aim to explore the optimal initial-final pair at one time slot using ML algorithm. Secondly, in order to find the optimal trajectories and initial-final positions pair for robots, an improved ML-based algorithm, i.e., D$^{3}$QN, is invoked for jointly planning trajectories and designing the beamforming. 

\vspace{-0.3cm}

\vspace{-0.4cm}
\subsection{LSTM-ARIMA Model for Prediction}
As a variant of recurrent neural network (RNN), the LSTM network has an excellent performance in efficiently handling non-stationary and non-linear sequences of data. But for long sequences, LSTM cannot completely solve the problem of gradient vanishment. The ARIMA model is not puzzled by this problem and provides a valid solution for a linear sequence of data. However, it is a time prediction model which essentially captures linear relationships, while nonlinear relationships cannot be involved. Therefore, we consider combining the advantages of the two models and propose a novel LSTM-ARIMA model for prediction.

\subsubsection{Data Pre-processing}
\vspace{-0.3cm}
\begin{figure}[htbp]
  \setlength{\belowcaptionskip}{-0.3cm} 
  \centering
  \includegraphics[scale = 0.09]{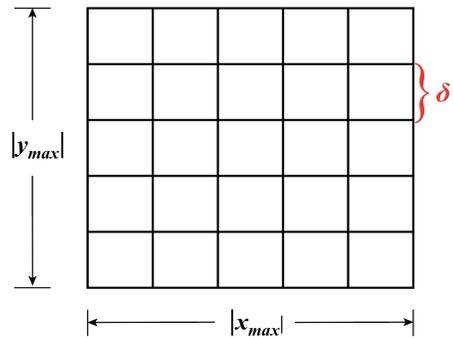}
  \caption{Discretization of the geographic map.}
  \label{disc}
\end{figure}
In order to characterize the geographic map of the environment, as shown in Fig.~\ref{disc}, the map is discretized into \emph{$(x_{max}y_{max})/\delta^{2}$} cells, where \emph{$x_{max}$}, \emph{$y_{max}$} and \emph{$\delta$} denote maximum bound of x-axis, maximum bound of y-axis and resolution of the map. To simplify the problem, we assume that \emph{$\delta$} is small enough which makes the size of cells can be approximated to the center of cells. Randomly generate \cite{IEEEhowto:Eckhardt} the position pairs \emph{$\mathbf{S}_{o} = \{S_{1},\cdots,S_{4N},S_{4N+1},\cdots,S_{5N},S_{5N+1},\cdots,S_{6N}\}$} as the historical positions data set, where \emph{$\mathbf{S}_{train} = \{S_{1},S_{2},\cdots,S_{4N}\}$} is selected as the training samples and \emph{$\mathbf{S}_{test} = \{S_{4N+1},S_{4N+2},\cdots,S_{5N}\}$} as the test samples for LSTM and ARIMA models. In order to reduce the influence of value range on network performance, it is necessary to normalize the values of training samples to [0,1]. By invoking Min-Max normalization method \cite{IEEEhowto:Patro}, the normalized samples can be expressed as
\par
\vspace{-0.2cm}
\noindent
\begin{align}\label{9}
  \tilde{S}_{\overline{N}, \overline{n}} = \frac{S_{\overline{N}, \overline{n}}-S_{min}}{S_{max}-S_{min}}, \hspace{0.5em} \overline{n} = 1, 2, \hspace{0.5em} \overline{N} \in \{1, 2, \cdots, 5N\}
\end{align}
\par
\vspace{-0.2cm}
\noindent
where \emph{$\tilde{S}_{\overline{N}, \overline{n}}$}, \emph{$S_{\overline{N}, \overline{n}}$}, \emph{$S_{max}$} and \emph{$S_{min}$} are the normalized values of positions pairs, the original values of positions pairs, the maximum values of all positions pairs, and the minimum values of all positions pairs, respectively. Note that, \emph{$\overline{n} = 1$} and \emph{$\overline{n} = 2$} represent the initial position and final position, while \emph{$S_{max}$} and \emph{$S_{min}$} are determined by all initial and final positions.

\subsubsection{Prediction Based on LSTM Model}
We take the normalized data \emph{$\tilde{\mathbf{S}} = \{\tilde{S}_{1},\tilde{S}_{2},\cdots,\tilde{S}_{4N}\}$} and denote \emph{$\overline{\mathbf{S}}^{\rm LSTM} = \{\overline{S}_{1}^{\rm LSTM},\overline{S}_{2}^{\rm LSTM},\cdots,\overline{S}_{N}^{\rm LSTM}\}$} as the input position pairs and predicted position pairs. According to the LSTM model, denote \emph{$\nu_{i}$}, \emph{$\nu_{f}$}, \emph{$\nu_{o}$}, \emph{H}, and \emph{U} as input gate, forget gate, output gate, size of input layer, and size of hidden memory, respectively. Thus, at each timestep \emph{t}, the definition of those parameters can be calculated as
\par
\vspace{-0.2cm}
\noindent
\begin{align}\label{9a}
  \mathbf{\nu}_{i}^{t} = g(\sum_{h} \omega_{i,h}\tilde{\mathbf{S}}_{i}^{t} + \sum_{u} \omega_{i,u}\mathbf{c}_{i}^{t-1}\mathbf{b}_{u}^{t-1}),
\end{align}
\par
\vspace{-0.2cm}
\noindent
\begin{align}\label{10}
  \mathbf{\nu}_{f}^{t} = g(\sum_{h} \omega_{f,h}\tilde{\mathbf{S}}_{f}^{t} + \sum_{u} \omega_{f,u}\mathbf{c}_{f}^{t-1}\mathbf{b}_{u}^{t-1}),
\end{align}
\par
\vspace{-0.2cm}
\noindent
\begin{align}\label{11}
  \mathbf{\nu}_{o}^{t} = g(\sum_{h} \omega_{o,h}\tilde{\mathbf{S}}_{o}^{t} + \sum_{u} \omega_{o,u}\mathbf{c}_{o}^{t-1}\mathbf{b}_{u}^{t-1}),
\end{align}
\noindent
where \emph{$g(\cdot)$}, \emph{$\mathbf{c}^{t-1}$} and \emph{$\mathbf{b}_{u}^{t}=\mathbf{\nu}_{f}^{t}\mathbf{b}_{u}^{t-1}+\mathbf{\nu}_{i}^{t}g(\sum_{h} \omega_{i,u}\tilde{\mathbf{S}}_{i}^{t} + \sum_{u} \omega_{i,u}\mathbf{c}_{i}^{t-1})$} denote the activation function, output of each hidden layer, state of each layer, respectively. Then, the predicted denomalized output \emph{$\overline{\mathbf{S}}^{\rm LSTM}$} of the model can be calculated by
\par
\vspace{-0.2cm}
\noindent
\begin{align}\label{12}
  \overline{S}_{n}^{\rm LSTM} = \mathbf{\nu}_{o}{\rm tanh}(\mathbf{b}_{u}^{t-1}) * (S_{max} - &S_{min}) + S_{min}, \nonumber \\
  &n \in \{1, 2, \cdots, N\}.
\end{align}
\par
\vspace{-0.2cm}
To prevent LSTM training from overfitting, the adjustment of weight on the output fully connected layer is increased by L2 normalization regularization \cite{IEEEhowto:Laarhoven}.

\subsubsection{Prediction Based on ARIMA Model}
ARIMA is an effective statistical model for time series analysis and forecasting, which is integrated of auto regression (AR) model, moving average (MA) model and difference model. The AR model and MA model can be expressed as
\par
\vspace{-0.2cm}
\noindent
\begin{align}\label{13}
  \overline{S}_{n}^{AR} = \mu_{n} + \sum\limits_{z=1}^{a}\kappa_{z}\tilde{S}_{n-z}, \hspace{0.5em} n \in \{1, 2, \cdots, N\},
\end{align}
\par
\vspace{-0.4cm}
\noindent
\begin{align}\label{14}
  \overline{S}_{n}^{MA} = \zeta_{n} + \sum\limits_{z=1}^{b}\varphi_{z}\zeta_{n-z} = \sum\limits_{z=1}^{b}\varphi_{z}L\zeta_{n}, \hspace{0.5em} n \in \{1, 2, \cdots, N\},
\end{align}
\par
\vspace{-0.2cm}
\noindent
where \emph{$\mu_{n}$}, \emph{a}, \emph{$\kappa_{z}$}, \emph{$\zeta_{n}$}, \emph{b}, \emph{$\varphi_{z}$}, and \emph{L} denote the constant value, order of autoregressive model, autoregressive coefficient, model errors, order of the moving average model, moving average coefficient, and lag operator, respectively. Then, in order to gurantee the stability of input data, a difference operator \emph{$\Delta^{d}$} is employed, which can be given by
\par
\vspace{-0.2cm}
\noindent
\begin{align}\label{d}
  \Delta^{d}\overline{S}_{n} = (1-L)^{d}(\overline{S}_{n}^{AR}+\overline{S}_{n}^{MA}), \hspace{0.5em} n \in \{1, 2, \cdots, N\}.
\end{align}
\par
\vspace{-0.2cm}
Thus, the the predicted denomalized output \emph{$\overline{\mathbf{S}}^{\rm ARIMA}$} of ARIMA model can be expressed as
\par
\vspace{-0.2cm}
\noindent
\begin{align}\label{ARIMA}
  &\overline{S}_{n}^{\rm ARIMA} = [\sum\limits_{z=1}^{a}\kappa_{z}(1-L)^{d}(\overline{S}_{n}^{AR}+\overline{S}_{n}^{MA}) + \sum\limits_{z=1}^{b}\varphi_{z}\zeta_{n-z} + \nonumber \\
  &\hspace{1em} \mu_{n} + \zeta_{n}] * (S_{max} - S_{min}) + S_{min}, \hspace{0.5em} n \in \{1, 2, \cdots, N\}.
\end{align}
\par
\vspace{-0.2cm}

\subsubsection{Fusion Based on Critic Weight Method}
In order to further improve the performance of the model, the critic weight method \cite{IEEEhowto:Diakoulaki} is exploited to assign weights to the predicted values of the two models. An evaluate indicator \emph{$\mathcal{I}$} is proposed to calculate the RMSE of the two models, which can be expressed as
\par
\vspace{-0.2cm}
\noindent
\begin{align}\label{Test}
  \mathcal{I} = \sqrt{\frac{1}{N}\sum\limits_{n=1}^{N}(\mid\overline{S}_{n}-S_{4N+n}\mid)}.
\end{align}
\par
\vspace{-0.2cm}
When \emph{$\mathcal{I}$} belows the threshold \emph{$\varsigma$}, the trained model can be accepted, the predicted data \emph{$\overline{\mathbf{S}} = \{\overline{S}_{1},\overline{S}_{2},\cdots,\overline{S}_{N}\}$} by LSTM model and ARIMA model are obtained by inputing \emph{$\mathbf{S}_{ini} = \{S_{5N+1},S_{5N+2},\cdots,S_{6N}\}$}. Accordingly, the predicted data \emph{$\overline{\mathbf{S}}$} after model fusion can be calculated as
\par
\vspace{-0.2cm}
\noindent
\begin{align}\label{F}
  \overline{S}_{n} = \overline{w}_{n}\overline{S}_{n}^{\rm LSTM} + \tilde{w}_{n}\overline{S}_{n}^{\rm ARIMA}, \hspace{0.5em} n \in \{1, 2, \cdots, N\}.
\end{align}
\par
\vspace{-0.2cm}
When weights \emph{$\overline{w}_{n}$} and \emph{$\tilde{w}_{n}$} are determined, the channel properties set $\mathbf{Z}$, initial position \emph{$\varepsilon_{i}$} and final position \emph{$\xi_{i}$} can be opted from the prediction pairs \emph{$\overline{\mathbf{S}}_{ini}$}. The detailed pseudo code is shown in \textbf{Algorithm~\ref{LSTM_ARIMA}}. The problem \eqref{8} can be reformulated as
\par
\vspace{-0.2cm}
\noindent
\begin{align}
  \max_{\mathbf{\upsilon}^{d},\Omega,\{p_{i}\},\varepsilon_{i}, \xi_{i}, \mathbf{Q}^{d}} \hspace*{1em}& \sum\limits_{i=1}^{\mathcal{X}} R_{i}(\mathbf{Q}_{i}^{d}) \label{8r}\\
  {\rm s.t.} \hspace*{1em}& R_{i}(q_{i}^{(\mathfrak{n})}) \geq \overline{R}, \hspace*{0.5em} \forall \mathfrak{n} \in \{0, 1, \cdots, N \}, \tag{\ref{8r}{a}} \label{8ra}\\
  & |\upsilon_{m}^{(\mathfrak{n})}| = 1, \hspace*{0.5em} \forall m \in \{1, 2, \cdots, M\}, \nonumber \\
  &\hspace{5em} \forall \mathfrak{n} \in \{0, 1, \cdots, N \}, \tag{\ref{8r}{b}} \label{8rb}\\
  & q_{i}^{(0)} = \varepsilon_{i}, q_{i}^{(N)} = \xi_{i}, \hspace*{0.5em} \forall \varepsilon_{i}, \xi_{i} \in \overline{\mathbf{S}}_{ini}, \tag{\ref{8r}{c}} \label{8rc} \\
  & |\dot{q}_{i}^{(\mathfrak{n})}| = V, \hspace*{0.5em} \forall \mathfrak{n} \in \{0, 1, \cdots, N \}, \tag{\ref{8r}{d}} \label{8rd} \\
  & q_{i}^{(\mathfrak{n})} \in \mathbf{Q}_{i}^{d}, \hspace*{0.5em} \forall \mathfrak{n} \in \{0, 1, \cdots, N \}, \tag{\ref{8r}{e}} \label{8re} \\
  & \eqref{8c} - \eqref{8e},\tag{\ref{8r}{f}} \label{8rf}
\end{align}
\par
\vspace{-0.2cm}
\noindent
where \emph{$\mathbf{Q}_{i}^{d}$}, \emph{$q_{i}^{(n)}$}, \emph{$\upsilon_{m}^{(n)}$}, and \emph{$N$} denote the trajectory based on discreted map, position in the trajectory, RIS constraint on each position, and total number of positions in each trajectory, respectively.

\begin{algorithm}[htbp]
  \setstretch{1}
  \caption{LSTM-ARIMA algorithm for prediction}
  \label{LSTM_ARIMA}
  \begin{algorithmic}[1]
  \REQUIRE ~~\\
  LSTM network structure, ARIMA model structure.\\
  \ENSURE The channel properties $\mathbf{Z}$, The predicted position pairs \emph{$\overline{\mathbf{S}}_{ini}$}.\\
  \STATE \textbf{Initialize:} Parameters of LSTM network, Parameters of ARIMA algorithm, $\mathcal{X}$ robots, geographic map.
  \STATE Randomly generate \emph{$\mathbf{S} = \{S_{1},S_{2},\cdots,S_{4N},S_{4N+1}\cdots,S_{5N}\}$} as historical data for robots.
  \STATE Split $\mathbf{S}$ into \emph{$\mathbf{S}_{train} = \{S_{1},S_{2},\cdots,S_{4N}\}$} and \emph{$\mathbf{S}_{test} = \{S_{4N+1},S_{4N+2},\cdots,S_{5N}\}$}.
  \STATE Normalize historical data by invoking Min-Max normalization method: $\tilde{S}_{\overline{N}, \overline{n}} = \frac{S_{\overline{N}, \overline{n}}-S_{min}}{S_{max}-S_{min}}$. 
  \STATE Input normalized \emph{$\tilde{\mathbf{S}} = \{\tilde{S}_{1},\tilde{S}_{2},\cdots,\tilde{S}_{4N}\}$} and \emph{$\tilde{\mathbf{S}} = \{\tilde{S}_{4N+1},\tilde{S}_{4N+2},\cdots,\tilde{S}_{5N}\}$}.
  \STATE Calculate \emph{$\nu_{i}$}, \emph{$\nu_{f}$}, \emph{$\nu_{o}$} of LSTM network.
  \FOR {n = 1 to N}
    \STATE Predict \emph{$\overline{S}_{n}^{\rm LSTM} = \mathbf{\nu}_{o}{\rm tanh}(\mathbf{b}_{u}^{t-1}) * (S_{max} - S_{min}) + S_{min}, \hspace{0.5em} n \in \{1, 2, \cdots, N\}$}.
    \STATE Calculate \emph{$\Delta^{d}\overline{S}_{n} = (1-L)^{d}(\overline{S}_{n}^{AR}+\overline{S}_{n}^{MA})$}.
    \STATE Predict \emph{$\overline{S}_{n}^{\rm ARIMA} = [\sum_{z=1}^{a}\kappa_{z}(1-L)^{d}(\overline{S}_{n}^{AR}+\overline{S}_{n}^{MA}) + \sum_{z=1}^{b}\varphi_{z}\zeta_{n-z} + \mu_{n} + \zeta_{n}] * (S_{max} - S_{min}) + S_{min}$}.
    \STATE Calculate \emph{$\overline{w}_{n}$} and \emph{$\tilde{w}_{n}$} by employing critic weight method.
    \STATE Calculate \emph{$\overline{S}_{n} = \overline{w}_{n}\overline{S}_{n}^{\rm LSTM} + \tilde{w}_{n}\overline{S}_{n}^{\rm ARIMA}$}.
  \ENDFOR
  \STATE Calculate \emph{$\mathcal{I} = \sqrt{\frac{1}{N}\sum\limits_{n=1}^{N}(\mid\overline{S}_{n}-S_{4N+n}\mid)}$}.
  \IF{\emph{$\mathcal{I} > \varsigma$}}
    \STATE Go to \textbf{line 1}
  \ENDIF
  \IF{\emph{$\mathcal{I} <= \varsigma$}}
    \STATE Accept trained LSTM-ARIMA fusion model.
    \STATE Input \emph{$\mathbf{S}_{ini}$} to the model and obtain $\mathbf{Z}$ and \emph{$\overline{\mathbf{S}}_{ini}$}.
  \ENDIF
  \end{algorithmic}
\end{algorithm}
\vspace{-0.2cm}
\begin{remark}\label{remark 1}
  The position set predicted by LSTM model constitutes an unstable sequence, and ARIMA model is able to make d-th order difference to the unstable sequence through its difference function, turning the sequence into a stable sequence.
\end{remark}

\vspace{-0.2cm}
\subsubsection{Complexity for LSTM-ARIMA Model}
The complexity of the proposed LSTM-ARIMA model is mainly related to the LSTM model, ARIMA model and the learning process. LSTM model maintains a total of four parameters, which are input gate \emph{$\nu_{i}$}, output gate \emph{$\nu_{o}$}, forget gate \emph{$\nu_{f }$} and candidate states\emph{$\nu_{c}=\mathbf{\nu}_{o}{\rm tanh}(\mathbf{b}_{u}^{t-1})$ }. The total size of parameters are 4len(\emph{$\nu_{f}$})(len(\emph{$\nu_{i}$}) + len(\emph{$ \nu_{c}$}) + len(\emph{$\nu_{o}$})), where the length of \emph{$\nu_{o}$} is the same with \emph {$\nu_{f}$}, \emph{$\nu_{c}$} is determined by \emph{$\nu_{i}$}, \emph{$\nu_{f}$} and \emph{$\nu_{o}$}, where len($\cdot$) denotes the length of parameter. Thus, the computation complexity of LSTM model is $4(\tilde{n}\tilde{m}+ \tilde{n}^{2} +\tilde{n})$ , where \emph{$\tilde{n}$} = len(\emph{$\nu_{f}$}), \emph{$\tilde{m}$} = len(\emph{$\nu_{i}$}). The computation of ARIMA model is determined by the length of sequence, which can be expressed as \emph{$\hat{N}$}. In terms of learning process, the computational complexity is related to the number of timesteps \emph{T} and episodes \emph{$N_{e}$}. Thus, the computation complexity can be expressed as \emph{$O$}(\emph{$4(\tilde{n}\tilde{m}+ \tilde{n}^{2} +\tilde{n})\hat{N}TN_{e}$}).

\vspace{-0.4cm}
\subsection{Dueling double deep Q-network Algorithm for Trajectories Planning and Passive Beamforming Design}
Our goal is to obtain the whole trajectory with a maximum sum-rate, while acquiring a local maximum sum-rate does not guarantee that the overall maximum sum-rate can be achieved. The double DQN algorithm \cite{IEEEhowto:Diakoulaki} improves the accuracy of the Q-value obtained by the DQN algorithm, which employs the current network to help estimate the Q-network. Specifically, it evaluates the action in the next state of the current state and chooses the optimal action, and then adopts it for the current state to estimate a new Q value. However, the optimal action selected and other unselected actions in the next state may have the same impact on the next state. On the basis of the double DQN algorithm, we introduce the dueling DQN \cite{IEEEhowto:Hasselt} algorithm to evaluate the advantages of actions selected in the next state. In this section, we propose an ML-based algorithm, namely, D$^{3}$QN-based algorithm, which is able to guide the robots to interact with the environment to train an optimal policy, and further for trajectory planning and the phase shifts of the RIS, as well as the power allocation from the AP to the robots. According to the Rician distribution and Shannon formula, the perfect channel estimation can be executed after the position, phase shift, and power allocation strategy are determined by the ML policy to obtain the reward.

\subsubsection{D$^{3}$QN-based Algorithm for Trajectories Planning and RIS Design}
In the D$^{3}$QN-based model, the AP acts as an agent, which is able to control both the power allocation policy from the AP to robots, the phase shift of RIS, and the robots' positions. At each timestep, the AP observes the state of the RIS-aided system and carries out an optimal action based on the policy, which is determined by Q-function. Following the selected action, the AP receives the reward when the current state of the model is transmitted to the next state, where the reward can be calculated by obtaining sum-rate from three robots.
\par
The state space \emph{$\mathbf{E} = \{e_{\overline{t}}\}$} at each epoch of the RIS-aided multi-robot networks is defined into three parts: the current phase shift \emph{$\{\theta_{m}\} \in [0,2\pi)$} of passive reflecting elements in the RIS at each position, the current position \emph{$q_{i}$} = (\emph{$x_{i}$},\emph{$y_{i}$},\emph{$h_{r}$}) of the robot \emph{$i$}, and the current group of allocation power \emph{$\{p_{i}\}$} from the AP to all the robots\footnote{The SIC decoding order is assumed perfect in this paper.}. Thus, the state space \emph{$\mathbf{E}$} can be expressed as
\par
\vspace{-0.2cm}
\noindent
\begin{align}\label{15}
  \mathbf{E} = 
  \begin{bmatrix} 
    \{\theta_{m}\} & \{q_{i}\} & \{p_{i}\}
  \end{bmatrix}.
\end{align}
\par
\vspace{-0.2cm}
The position \emph{$q_{i}$} on the trajectories for robot \emph{i} is determined by the previous position except initial-final positions generated from \emph{$\overline{S}_{ini}$}. The primary state space complexity is calculated as ($M+2\mathcal{X}$). Additionally, the total number of positions \emph{$N_{i}$} of each robot should meet the condition \emph{$N_{i} >= \max\limits_{i} (|x_{\varepsilon_{i}} - x_{\xi_{i}}| + |y_{\varepsilon_{i}} - y_{\xi_{i}}| - 1)$}, which ensures that all robots can reach the final position from the initial position. The action space \emph{$\mathbf{F} = \{f_{\overline{t}}\}$} at each epoch of the RIS-aided multi-robot networks is defined into three parts: the available quantity of phase shifts \emph{$\{\frac{2\pi n_{0}}{2^{B_{0}}}, n_{0} = 0,1,2,\cdots,2^{B_{0}}-1\}$}, the distance with moving direction \emph{$\mathbf{D}_{i} = \{d_{r},d_{l},d_{0},d_{u},d_{d}\}$} for robot \emph{i}, the available quantity of power allocation \emph{$\{p^{1},p^{2},\cdots,p^{a}\}_{i}$} for robot \emph{i}. \emph{$B_{0}$}, \emph{$d_{g}, \mathbf{g}=\{r,l,0,u,d\}$}, \emph{$a$} denote the resolution for the RIS phase shift, the right-left-stillness-up-down direction with 1 unit pace, and the total number of the available power allocated to the robots. Note that action ``stillness" in one state is only applicable to at most \emph{$\mathcal{X}-1$} robots to choice. Thus, the action state \emph{$\mathbf{F}$} can be expressed as
\par
\vspace{-0.2cm}
\noindent
\begin{align}\label{16}
  \mathbf{F} = 
  \begin{bmatrix} 
    \big \{\big \{\frac{2\pi n_{0}}{2^{B_{0}}} \big\}_{m} \big \} & \{\mathbf{D}_{i}\} & \{\{p^{1},p^{2},\cdots,p^{a}\}_{i}\}
  \end{bmatrix}
\end{align}
\par
\vspace{-0.2cm}
Accordingly, the primary action space complexity is calculated as ($2^{B_{0}} \cdot M+5\mathcal{X}+a\mathcal{X}$). The reward is a considerable factor in the optimization of trajectory planning and passive beamforming design, which can be determined by observing the different sum-rate of all robots between two adjacent positions. In general, the robots move from the current position to the next position at each epoch. However, in order to avoid collisions among robots, if the distance between any two robots equals $\delta$, the robots are regarded as located in adjacent cells, and there exists at least one action that is unable to be selected. According to the definition of reinforcement learning, the rewards brought by the unavailable action can be defined as -10. The reward function can be calculated as
\par
\vspace{-0.2cm}
\noindent
\begin{align}\label{17}
  \mathcal{R} =\left\{
      \begin{array}{lr}
       -10, \hspace{0.5em} \forall i_1,i_2 \in \{1, 2, \cdots, \mathcal{X}\}, ||q_{i_1} - q_{i_2}|| = \delta, & \\
       \Big [\sum\limits_{i=1}^{\mathcal{X}} R_{i}(q_{i}(\mathfrak{n}+1)) \Big ]_{(e)} - \Big [\sum\limits_{i=1}^{\mathcal{X}} R_{i}(q_{i}(\mathfrak{n})) \Big ]_{(e)}, \\
       \hspace{17em} \mathrm{Otherwise},
      \end{array}
    \right.
\end{align}
\par
\vspace{-0.2cm}
\noindent
where \emph{$[\cdot]_{e}$} denote the sum-rate of trajectories of all robots at \emph{e}-th epoch. Thus, it is observed from \eqref{17}, that maximizing the long-term sum rewards makes the dedication to maximizing the optimized trajectories and passive beamforming of the RIS-aided multi-robot system. According to the double DQN model, there are two networks with the same structure: the current Q-network and the target Q-network. The agent selects actions with the maximum Q value from the current Q-network, which can be expressed as
\par
\vspace{-0.2cm}
\noindent
\begin{align}\label{18}
  f_{\overline{t}}^{max}(e_{\overline{t}}^{'},\psi_{\overline{t}})={\rm arg} \max_{f_{\overline{t}}^{'}} Q[\Gamma(e_{\overline{t}}^{'}),f_{\overline{t}};\psi_{\overline{t}}],
\end{align}
\par
\vspace{-0.2cm}
\noindent
and the Q-value in the target Q-network can be calculated by
\begin{align}\label{19}
  &Q^{'}[\Gamma(e_{\overline{t}}^{'}),f_{\overline{t}}^{max}(e_{\overline{t}}^{'},\psi_{\overline{t}});\overline{\psi}_{\overline{t}}] = \nonumber \\
  &\hspace{7em}\mathcal{R} + \eta Q[\Gamma(e_{\overline{t}}^{'}),f_{\overline{t}}^{max}(e_{\overline{t}}^{'},\psi_{\overline{t}});\overline{\psi}_{\overline{t}}],
\end{align}
\par
\vspace{-0.2cm}
\noindent
where \emph{$\Gamma(\cdot)$}, \emph{$\psi_{\overline{t}}$}, and \emph{$\overline{\psi}_{\overline{t}}$} are the feature vector of state, the parameters for the current Q-network and target Q-network, respectively. The update method for \emph{$\psi_{\overline{t}}$} and \emph{$\overline{\psi}_{\overline{t}}$} are independent. The updated for \emph{$\psi_{t}$} can be expressed as
\par
\vspace{-0.2cm}
\noindent
\begin{align}\label{20}
  \psi_{t}^{'} &= \psi_{t} + \eta_{0}\{Q^{'}[\Gamma(e_{\overline{t}}^{'}),f_{\overline{t}}^{max}(e_{\overline{t}}^{'},\psi_{\overline{t}});\overline{\psi}_{\overline{t}}] \nonumber \\
  &\hspace{5em}- Q[\Gamma(e_{\overline{t}}),f_{\overline{t}};\psi_{\overline{t}}]\}\triangledown_{\psi_{t}}Q[\Gamma(e_{\overline{t}}),f_{\overline{t}};\psi_{\overline{t}}], \nonumber \\
  & = \psi_{t} + \eta_{0}\{\mathcal{R} + \eta Q[\Gamma(e_{\overline{t}}^{'}),f_{\overline{t}}^{max}(e_{\overline{t}}^{'},\psi_{\overline{t}});\overline{\psi}_{\overline{t}}] \nonumber \\
  &\hspace{5em}- Q[\Gamma(e_{\overline{t}}),f_{\overline{t}};\psi_{\overline{t}}]\}\triangledown_{\psi_{t}}Q[\Gamma(e_{\overline{t}}),f_{\overline{t}};\psi_{\overline{t}}],
\end{align}
\par
\vspace{-0.2cm}
\noindent
where the \emph{$\triangledown_{\{\cdot\}}$} is the gradient operator. The loss function can be expressed as follows:
\par
\vspace{-0.2cm}
\noindent
\begin{align}\label{loss for DDQN}
  {\rm Loss(\psi_{\overline{t}})} &= \mathbb{E}\{\{Q^{'}[\Gamma(e_{\overline{t}}^{'}),f_{\overline{t}}^{max}(e_{\overline{t}}^{'},\psi_{\overline{t}});\overline{\psi}_{\overline{t}}] \nonumber \\
  &\hspace{9em}- Q[\Gamma(e_{\overline{t}}),f_{\overline{t}};\psi_{\overline{t}}]\}^{2}\}, \nonumber \\
  &= \mathbb{E}\{\{\mathcal{R} + \eta Q[\Gamma(e_{\overline{t}}^{'}),f_{\overline{t}}^{max}(e_{\overline{t}}^{'},\psi_{\overline{t}});\overline{\psi}_{\overline{t}}] \nonumber \\
  &\hspace{9em}- Q[\Gamma(e_{\overline{t}}),f_{\overline{t}};\psi_{\overline{t}}]\}^{2}\}.
\end{align}
\par
\vspace{-0.2cm}
For \emph{$\overline{\psi}_{\overline{t}}$}, it can be updated every \emph{$\mathbb{T}$} epochs according to \emph{$\psi_{\overline{t}}$}, which can be expressed as
\par
\vspace{-0.2cm}
\noindent 
\begin{align}\label{update params}
  \overline{\psi}_{\overline{t}} = \psi_{\overline{t}} \big |_{\mathbb{T}}.
\end{align}
\par
\vspace{-0.2cm}
However, when evaluating the potential actions for the current positions, there are two or more actions that bring the same effect on the same states. In this case, the robot may sample the wrong action, which will affect the subsequent actions, resulting in the inability to finally obtain the optimal trajectory. In order to reduce the impact of this case, we employ the dueling architecture in the target Q-network to evaluate state and action, respectively. The target Q-value obtained by equation \eqref{19} can be rewritten as:
\par
\vspace{-0.2cm}
\noindent
\begin{align}\label{21}
  &Q^{'}[\Gamma(e_{\overline{t}}^{'}),f_{\overline{t}}^{max}(e_{\overline{t}}^{'},\psi_{\overline{t}});\overline{\psi}_{\overline{t}}^{\rm C}, \overline{\psi}_{\overline{t}}^{\rm D1}, \overline{\psi}_{\overline{t}}^{\rm D2}] \nonumber \\
  &= Q_{e}(\Gamma(e_{\overline{t}}^{'});\overline{\psi}_{\overline{t}}^{\rm C},\overline{\psi}_{\overline{t}}^{\rm D2}) + Q_{f}[\Gamma(e_{\overline{t}}^{'}),f_{\overline{t}}^{max}(e_{\overline{t}}^{'},\psi_{\overline{t}});\overline{\psi}_{\overline{t}}^{\rm C}, \overline{\psi}_{\overline{t}}^{\rm D1}],
\end{align}
\par
\vspace{-0.2cm}
\noindent
where the \emph{$Q_{e}(\cdot)$}, \emph{$Q_{f}(\cdot)$}, \emph{$\overline{\psi}_{\overline{t}}^{\rm C}$}, \emph{$\overline{\psi}_{\overline{t}}^{\rm D1}$}, and \emph{$\overline{\psi}_{\overline{t}}^{\rm D2}$} denote the state evaluate function, action evaluate function, parameters for convolutional layers, parameters for dense layer to present advantage function, and parameters for dense layer to present value function, respectively. \emph{$\overline{\psi}_{\overline{t}}$} denotes the parameters of double DQN network architecture, including three convolutional layers and two dense layers in series. In contrast, \emph{$\overline{\psi}_{\overline{t}}^{\rm C}$} denotes the parameters of same three convolutional layers in dueling DQN structure, while \emph{$\overline{\psi}_{\overline{t}}^{\rm D1}$} and \emph{$\overline{\psi}_{\overline{t}}^{\rm D2}$} indicate the parameters of two dense layers in parallel. Note that, we only consider to split state and action of target Q-network in double DQN networks. Additionally, prioritized experience replay (PER) has been employed in dueling DQN network architecture. In the experience pool, PER extracts samples with a large absolute value of \emph{$\big |Q^{'}[\Gamma(e_{\overline{t}}^{'}),f_{\overline{t}}^{max}(e_{\overline{t}}^{'},\psi_{\overline{t}});\overline{\psi}_{\overline{t}}]-Q^{'}[\Gamma(e_{\overline{t}}^{'}),f_{\overline{t}}^{max}(e_{\overline{t}}^{'},\psi_{\overline{t}});\overline{\psi}_{\overline{t}}^{\rm C}, \overline{\psi}_{\overline{t}}^{\rm D1}, \overline{\psi}_{\overline{t}}^{\rm D2}] \big |$} for training, which is capable of speeding up the learning of the Q function. More importantly, PER can enhance the identifiability of \emph{$Q_{e}(\cdot)$} and \emph{$Q_{f}(\cdot)$}, we introduce a loss function, where the target Q-value can be rewritten as
\par
\vspace{-0.2cm}
\noindent
\begin{align}\label{22}
  &Q^{'}[\Gamma(e_{\overline{t}}^{'}),f_{\overline{t}}^{max}(e_{\overline{t}}^{'},\psi_{\overline{t}});\overline{\psi}_{\overline{t}}^{\rm C}, \overline{\psi}_{\overline{t}}^{\rm D1}, \overline{\psi}_{\overline{t}}^{\rm D2}] \nonumber \\
  &= Q_{e}(\Gamma(e_{\overline{t}}^{'});\overline{\psi}_{\overline{t}}^{\rm C},\overline{\psi}_{\overline{t}}^{\rm D2}) + Q_{f}[\Gamma(e_{\overline{t}}^{'}),f_{\overline{t}}^{max}(e_{\overline{t}}^{'},\psi_{\overline{t}});\overline{\psi}_{\overline{t}}^{\rm C}, \overline{\psi}_{\overline{t}}^{\rm D1}] \nonumber \\
  &\hspace{1em}- \frac{1}{\mathbf{\mathcal{A}}}\sum_{(f_{e})^{'}\in \mathbf{\mathcal{A}}}Q_{f}[\Gamma(e_{\overline{t}}^{'}),(f_{e})^{'}(e_{\overline{t}}^{'},\psi_{\overline{t}});\overline{\psi}_{\overline{t}}^{\rm C}, \overline{\psi}_{\overline{t}}^{\rm D1}],
\end{align}
\par
\vspace{-0.2cm}
\noindent
where the \emph{$\mathbf{\mathcal{A}}$}, \emph{$(f_{e}^{max})^{'}(\cdot)$} are the action space of current Q-network and an sampled action from action space of current Q-network, respectively. The updated for \emph{$\{\overline{\psi}_{\overline{t}}^{\rm C}, \overline{\psi}_{\overline{t}}^{\rm D1}, \overline{\psi}_{\overline{t}}^{\rm D2}\}$} can be expressed as \eqref{20u}, and the loss function can be calculated as \eqref{23}.
\par
\vspace{-0.2cm}
\noindent
\begin{figure*}[htbp]
  \normalsize 
  \begin{align}\label{20u}
    \{\overline{\psi}_{\overline{t}}^{\rm C}, \overline{\psi}_{\overline{t}}^{\rm D1}, \overline{\psi}_{\overline{t}}^{\rm D2}\}^{'} &= \{\overline{\psi}_{\overline{t}}^{\rm C}, \overline{\psi}_{\overline{t}}^{\rm D1}, \overline{\psi}_{\overline{t}}^{\rm D2}\} + \eta_{0}\{Q^{'}[\Gamma(e_{\overline{t}}^{'}),f_{\overline{t}}^{max}(e_{\overline{t}}^{'},\psi_{\overline{t}});\overline{\psi}_{\overline{t}}^{\rm C}, \overline{\psi}_{\overline{t}}^{\rm D1}, \overline{\psi}_{\overline{t}}^{\rm D2}] - Q[\Gamma(e_{\overline{t}}),f_{\overline{t}};\psi_{\overline{t}}]\}\triangledown_{\psi_{t}}Q[\Gamma(e_{\overline{t}}),f_{\overline{t}};\psi_{\overline{t}}], \nonumber \\
    & = \{\overline{\psi}_{\overline{t}}^{\rm C}, \overline{\psi}_{\overline{t}}^{\rm D1}, \overline{\psi}_{\overline{t}}^{\rm D2}\} + \eta_{0}\{\mathcal{R} + \eta \{Q_{e}(\Gamma(e_{\overline{t}}^{'});\overline{\psi}_{\overline{t}}^{\rm C},\overline{\psi}_{\overline{t}}^{\rm D2}) + Q_{f}[\Gamma(e_{\overline{t}}^{'}),f_{\overline{t}}^{max}(e_{\overline{t}}^{'},\psi_{\overline{t}});\overline{\psi}_{\overline{t}}^{\rm C}, \overline{\psi}_{\overline{t}}^{\rm D1}] \nonumber \\
    &\hspace{5em}- \frac{1}{\mathbf{\mathcal{A}}}\sum_{(f_{e})^{'}\in \mathbf{\mathcal{A}}}Q_{f}[\Gamma(e_{\overline{t}}^{'}),(f_{e})^{'}(e_{\overline{t}}^{'},\psi_{\overline{t}});\overline{\psi}_{\overline{t}}^{\rm C}, \overline{\psi}_{\overline{t}}^{\rm D1}]\} - Q[\Gamma(e_{\overline{t}}),f_{\overline{t}};\psi_{\overline{t}}]\}\triangledown_{\psi_{t}}Q[\Gamma(e_{\overline{t}}),f_{\overline{t}};\psi_{\overline{t}}],
  \end{align}
  \hrulefill \vspace*{0pt}
\end{figure*}

\par
\vspace{-0.8cm}
\noindent
\begin{figure*}[htbp]
  \normalsize 
  \begin{align}\label{23}
  {\rm Loss(\{\overline{\psi}_{\overline{t}}^{\rm C}, \overline{\psi}_{\overline{t}}^{\rm D1}, \overline{\psi}_{\overline{t}}^{\rm D2}\})} = &\mathbb{E}\{\{Q_{e}(\Gamma(e_{\overline{t}}^{'});\overline{\psi}_{\overline{t}}^{\rm C},\overline{\psi}_{\overline{t}}^{\rm D2}) + Q_{f}[\Gamma(e_{\overline{t}}^{'}),f_{\overline{t}}^{max}(e_{\overline{t}}^{'},\psi_{\overline{t}});\overline{\psi}_{\overline{t}}^{\rm C}, \overline{\psi}_{\overline{t}}^{\rm D1}] \nonumber \\
  &\hspace{8em}- \frac{1}{\mathbf{\mathcal{A}}}\sum_{(f_{e})^{'}\in \mathbf{\mathcal{A}}}Q_{f}[\Gamma(e_{\overline{t}}^{'}),(f_{e})^{'}(e_{\overline{t}}^{'},\psi_{\overline{t}});\overline{\psi}_{\overline{t}}^{\rm C}, \overline{\psi}_{\overline{t}}^{\rm D1}] - Q[\Gamma(e_{\overline{t}}),f_{\overline{t}};\psi_{\overline{t}}]\}^{2}\}.
  \end{align}
  \hrulefill \vspace*{0pt}
\end{figure*}
\par
\noindent

\begin{remark}\label{remark 2}
  The advantage function refers to the advantage of the action value function compared to the value function of the current state. If the advantage function is greater than 0, it means that the action is worse than the average action, otherwise the current action is not as good as the average action. To avoid situations where the advantage function is equal to 0, we need to centralize the advantage function.
\end{remark}
\vspace{-0.2cm}
Thus, according to the D$^{3}$QN model mentioned above, through interaction with the environment, the agent is able to find the optimal policy for the robot trajectories planning and phase designing in the RIS. The detailed pseudo code is shown in \textbf{Algorithm~\ref{DDDQN}}.
\begin{figure*}[htbp]
\begin{breakablealgorithm}
  \caption{${\rm D}^{3}$QN algorithm for trajectories planning and beamforming design}\label{DDDQN}
  \begin{algorithmic}[1]
  \setstretch{1}
  \REQUIRE ~~\\
  DQN network structure, LSTM network structure, ARIMA structure.
  \ENSURE Parameters of ${\rm D}^{3}$QN network, optimal initial-final pair $\overline{\mathbf{S}}_{\mathrm{op}}^{'} = \{\varepsilon_{\mathrm{op}}^{'}, \xi_{\mathrm{op}}^{'}\}$.
  \STATE \textbf{Initialize:} 2D moving space $\mathbf{M}$ for robots, reply memory $\mathbf{\mathcal{D}}$, the number of episodes $E$, minimal batch size $\mathcal{M}$, episodes indicator $\tilde{T}$, parameters update frequency for target Q network $\mathbb{T} < \tilde{T}$, temporary reward buffer vector $\mathbf{\mathfrak{R}} = \{\mathfrak{R}_{1}, \mathfrak{R}_{2}, \cdots, \mathfrak{R}_{N}\}$, and temporary initial-final position pairs buffer vector $\overline{\mathbf{S}}^{'} = \{\overline{\mathbf{S}}_{1}^{'}, \overline{\mathbf{S}}_{2}^{'},\cdots,\overline{\mathbf{S}}_{N}^{'}\}$, $\eta$, $\epsilon$, $\mathbf{Q}$, $\mathbf{Q}^{'}$, $\mathcal{R}$, $\mathbf{E}$, $\mathbf{F}$, $\psi_{\overline{t}}$, $\overline{\psi}_{\overline{t}}$, $\overline{\psi}_{\overline{t}}^{\rm C}$, $\overline{\psi}_{\overline{t}}^{\rm D1}$, and $\overline{\psi}_{\overline{t}}^{\rm D2}$, $\mathbf{\Psi}$, $\{p_{i}\}$, $\{\{p^{1},p^{2},\cdots,p^{a}\}_{i}\}$, $N$.
  \STATE Explore the positions of AP, RIS, and boundaries in $\mathbf{M}$.
  \STATE Train LSTM-ARIMA model for initial-final position pairs prediction.
  \STATE Generate $\{\varepsilon_{i}^{'}$, $\xi_{i}^{'}\}$ by $\overline{\mathbf{S}}_{ini}$ for all the robots.
  \FOR {episode $\tilde{t}$ from 1 to \emph{$\overline{T}$}}
  \STATE The agent randomly selects $e_{\overline{t}} \in \mathbf{E}$ and $\Gamma(e_{\overline{t}})$.
  \STATE Input $\Gamma(e_{\overline{t}})$ to current Q-network and obtain Q-values with all actions. 
  \STATE Sample $f_{\overline{t}}$ of each $e_{\overline{t}}$ by invoking $\epsilon$-greedy policy.
  \STATE Determine the decoding order based on the current state by NOMA method.
  \STATE Execute $f_{\overline{t}}$ and obtain $\Gamma(e_{\overline{t}}^{'})$ of new state, observe $\mathcal{R}$.
    \IF {The reward achieve the terminated condition $|\mathcal{R} - \mathcal{R}_{0}| \leq \hat{\mathcal{R}}$}
      \STATE Update for D$^{3}$QN-network at current episode is terminal, $\textbf{is\_terminated = true}$.
    \ENDIF
  \STATE Store transition ($\Gamma(e_{\overline{t}})$, $f_{\overline{t}}$, $\mathcal{R}$, $\Gamma(e_{\overline{t}}^{'})$, \textbf{is\_terminated}) in $\mathbf{\mathcal{D}}$.
  \STATE Update $e_{\overline{t}}^{'}$ as $e_{\overline{t}}$.
  \STATE Sample $\mathcal{M}$ of transition ($\Gamma(e_{\overline{t}})^{l}$, $f_{\overline{t}}^{l}$, $\mathcal{R}^{l}$, $\Gamma(e_{\overline{t}}^{'})^{l}$, $\textbf{is\_terminated}^{l}$) from $\mathbf{\mathcal{D}}$, $l = 1, 2, 3, \cdots, \mathcal{M}$. 
  \STATE Calculate $Q^{'}[\Gamma(e_{\overline{t}}^{'}),f_{\overline{t}}^{max}(e_{\overline{t}}^{'},\psi_{\overline{t}});\overline{\psi}_{\overline{t}}^{\rm C}, \overline{\psi}_{\overline{t}}^{\rm D1}, \overline{\psi}_{\overline{t}}^{\rm D2}] = $ \\
  $\left\{ \begin{array}{lr} \mathcal{R}, \hspace{2em} {\rm if\ \textbf{is}\textbf{\_terminated}\ is\ true}, & \\ Q_{e}(\Gamma(e_{\overline{t}}^{'});\overline{\psi}_{\overline{t}}^{\rm C},\overline{\psi}_{\overline{t}}^{\rm D2}) + \{Q_{f}[\Gamma(e_{\overline{t}}^{'}),f_{\overline{t}}^{max}(e_{\overline{t}}^{'},\psi_{\overline{t}});\overline{\psi}_{\overline{t}}^{\rm C}, \overline{\psi}_{\overline{t}}^{\rm D1}] - \frac{1}{\mathbf{\mathcal{A}}}\sum_{(f_{e})^{'}\in \mathbf{\mathcal{A}}}Q_{f}[\Gamma(e_{\overline{t}}^{'}), (f_{e})^{'}(e_{\overline{t}}^{'},\psi_{\overline{t}}); \overline{\psi}_{\overline{t}}^{\rm C}, \overline{\psi}_{\overline{t}}^{\rm D1}]\}, \\
  &\hspace{-9em} {\rm if\ \textbf{is}\textbf{\_terminated}\ is\ false}, \end{array}\right.$ 
  \STATE Perform a gradient descent step to calculate all parameters of target Q-network: \\
  $\{\overline{\psi}_{\overline{t}}^{\rm C}, \overline{\psi}_{\overline{t}}^{\rm D1},\overline{\psi}_{\overline{t}}^{\rm D2}\}^{'} = \{\overline{\psi}_{\overline{t}}^{\rm C}, \overline{\psi}_{\overline{t}}^{\rm D1}, \overline{\psi}_{\overline{t}}^{\rm D2}\} + \eta_{0}\{\mathcal{R} + \eta \{Q_{e}(\Gamma(e_{\overline{t}}^{'});\overline{\psi}_{\overline{t}}^{\rm C},\overline{\psi}_{\overline{t}}^{\rm D2}) + Q_{f}[\Gamma(e_{\overline{t}}^{'}), f_{\overline{t}}^{max}(e_{\overline{t}}^{'},\psi_{\overline{t}});\overline{\psi}_{\overline{t}}^{\rm C}, \overline{\psi}_{\overline{t}}^{\rm D1}] $\\ \hspace{8.5em}$- \frac{1}{\mathbf{\mathcal{A}}}\sum_{(f_{e})^{'}\in \mathbf{\mathcal{A}}}Q_{f}[\Gamma(e_{\overline{t}}^{'}),(f_{e})^{'}(e_{\overline{t}}^{'},\psi_{\overline{t}});\overline{\psi}_{\overline{t}}^{\rm C}, \overline{\psi}_{\overline{t}}^{\rm D1}]\} - Q[\Gamma(e_{\overline{t}}),f_{\overline{t}};\psi_{\overline{t}}]\}\triangledown_{\psi_{t}}Q[\Gamma(e_{\overline{t}}),f_{\overline{t}};\psi_{\overline{t}}]$.
  \IF {$\overline{T} \% \mathbb{T} = 1$}
    \STATE Update parameters $\{\overline{\psi}_{\overline{t}}^{\rm C}, \overline{\psi}_{\overline{t}}^{\rm D1}, \overline{\psi}_{\overline{t}}^{\rm D2}\} = \{\overline{\psi}_{\overline{t}}^{\rm C}, \overline{\psi}_{\overline{t}}^{\rm D1}, \overline{\psi}_{\overline{t}}^{\rm D2}\}^{'}$
  \ENDIF
  \IF {\{$\tilde{t} = \overline{T}$\} or \{$\mathcal{R}$ remains the same for $\tilde{T}$ consecutive episodes\}}
    \STATE Update for D$^{3}$Q network is end, reset $\mathbf{E}$, $\mathbf{F}$.
    \STATE Execute $\mathfrak{R}_{\tilde{t}} = \mathcal{R}$, and $\overline{\mathbf{S}}_{\tilde{t}}^{'} = \{\varepsilon_{i}^{'}, \xi_{i}^{'}\}$.
  \ENDIF
  \ENDFOR
  \STATE Obtain $\overline{\mathbf{S}}_{\mathrm{op}}^{'} = \{\varepsilon_{\mathrm{op}}^{'}, \xi_{\mathrm{op}}^{'}\}$ from $N$ explored pairs according to $\mathfrak{R}$.
  \end{algorithmic}
\end{breakablealgorithm}
\end{figure*}
\par
\vspace{-0.2cm}
\noindent
\begin{remark}\label{remark 3}
  The optimal policy of the Agent is always to choose the best action in any given state, while the best action often has smaller Q-values than the non-optimal ones in most cases. Such a problem is called the over-estimations of action value (Q-value). To overcome this problem, all the optional actions based on the current state need to be additionally evaluated as much as possible.
\end{remark}
\vspace{-0.2cm}

\subsubsection{Dueling DQN-based Algorithm for Trajectories Planning and RIS Design}
According to the dueling DQN-based algorithm, the AP acts as an agent, where the equipped controller has the capability of determining power allocation policy from the AP to robots, the phase shift of RIS, and the robots' positions. The state space \emph{$\mathbf{E} = \{e_{\overline{t}}\}$}, action space \emph{$\mathbf{F} = \{f_{\overline{t}}\}$}, and reward function \emph{$\mathcal{R}$} in double DQN-based algorithm are the same with D$^{3}$QN-based algorithm. According to dueling DQN architecture, the Q-function is split as follows:
\par
\vspace{-0.2cm}
\noindent
\begin{align}\label{25}
  &Q^{'}[\Gamma(e_{\overline{t}}^{'}),f_{\overline{t}}(e_{\overline{t}}^{'},\psi_{\overline{t}});\overline{\psi}_{\overline{t}}^{\rm C}, \overline{\psi}_{\overline{t}}^{\rm D1}, \overline{\psi}_{\overline{t}}^{\rm D2}] \nonumber \\
  &= Q_{e}(\Gamma(e_{\overline{t}}^{'});\overline{\psi}_{\overline{t}}^{\rm C},\overline{\psi}_{\overline{t}}^{\rm D2}) + Q_{f}[\Gamma(e_{\overline{t}}^{'}),f_{\overline{t}}(e_{\overline{t}}^{'},\psi_{\overline{t}});\overline{\psi}_{\overline{t}}^{\rm C}, \overline{\psi}_{\overline{t}}^{\rm D1}] \nonumber \\
  &\hspace{1em} - \frac{1}{\mathbf{\mathcal{A}}}\sum_{(f_{e})^{'}\in \mathbf{\mathcal{A}}}Q_{f}[\Gamma(e_{\overline{t}}^{'}),(f_{e})^{'}(e_{\overline{t}}^{'},\psi_{\overline{t}});\overline{\psi}_{\overline{t}}^{\rm C}, \overline{\psi}_{\overline{t}}^{\rm D1}],
\end{align}
\par
\vspace{-0.2cm}
\noindent
and the loss function can be expressed as
\par
\vspace{-0.2cm}
\noindent
\begin{align}\label{26}
  &{\rm Loss(\{\overline{\psi}_{\overline{t}}^{\rm C}, \overline{\psi}_{\overline{t}}^{\rm D1}, \overline{\psi}_{\overline{t}}^{\rm D2}\})} \nonumber \\
  &\hspace{2em}= \mathbb{E}\{\{Q_{e}(\Gamma(e_{\overline{t}}^{'});\overline{\psi}_{\overline{t}}^{\rm C},\overline{\psi}_{\overline{t}}^{\rm D2})  \nonumber \\
  &\hspace{3em}+ Q_{f}[\Gamma(e_{\overline{t}}^{'}),f_{\overline{t}}(e_{\overline{t}}^{'},\psi_{\overline{t}});\overline{\psi}_{\overline{t}}^{\rm C}, \overline{\psi}_{\overline{t}}^{\rm D1}] \nonumber \\
  &\hspace{3em}- \frac{1}{\mathbf{\mathcal{A}}}\sum_{(f_{e})^{'}\in \mathbf{\mathcal{A}}}Q_{f}[\Gamma(e_{\overline{t}}^{'}),(f_{e})^{'}(e_{\overline{t}}^{'},\psi_{\overline{t}});\overline{\psi}_{\overline{t}}^{\rm C}, \overline{\psi}_{\overline{t}}^{\rm D1}] \nonumber \\ 
  &\hspace{3em}- Q[\Gamma(e_{\overline{t}}),f_{\overline{t}};\psi_{\overline{t}}]\}^{2}\}.
\end{align}
\par
\vspace{-0.2cm}

\subsubsection{Double DQN-based Algorithm for Trajectories Planning and RIS Design}
According to double DQN-based algorithm, the AP acts as an agent, where the power allocation policy from the AP to robots, phase shift of RIS, and the robots' positions are determined by the controller. The state space \emph{$\mathbf{E} = \{e_{\overline{t}}\}$}, action space \emph{$\mathbf{F} = \{f_{\overline{t}}\}$}, and reward function \emph{$\mathcal{R}$} in double DQN-based algorithm are the same with D$^{3}$QN-based algorithm, while the Q-function and loss function follows \eqref{19} and \eqref{loss for DDQN}. Moreover, the parameters of target Q-network are updated follows \eqref{update params}.

\subsubsection{Complexity for D$^{3}$QN Algorithm}
The complexity of performance of the D$^{3}$QN algorithm depends on the convolution layers and learning process. Thus, the complexity of convolution layers can be expressed as \emph{$\sum_{\overline{d}=1}^{\overline{D}} \overline{M}_{\overline{d}}^{2}\overline{K}_{\overline{d}}^{2}\overline{C}_{\overline{d}-1}\overline{C}_{\overline{d}}$}, where \emph{$\overline{D}$}, \emph{$\overline{d}$}, \emph{$\overline{K}$}, and \emph{$\overline{C}_{d}$} denote the total number of convolution layers, the number of rows of feature two-dimension data, length of convolution kernel, and the number of convolution kernels of \emph{$\overline{d}$}-th layer, respectively. For learning process, according to the reinforcement learning method, denote \emph{$\overline{F}$}, \emph{$\overline{E}$} and \emph{$\hat{T}$} as the total number of actions that the agent is able to choose, the number of saved state-action pairs, and the number of timesteps. Thus, the computation complexity of D$^{3}$QN can be expressed as \emph{$O$}(\emph{$(\sum_{\overline{d}=1}^{\overline{D}} \overline{M}_{d}^{2}\overline{K}_{d}^{2}\overline{C}_{d-1}\overline{C}_{d}+|\overline{F}|+\overline{E})\hat{T}$}).

\vspace{-0.4cm}
\section{Numerical Results}
In this section, we provide simulation results to verify the effectiveness of the proposed machine learning-based optimization algorithms for joint trajectories planning and passive beamforming design, as well as the performance of the algorithms. In the simulations, the number of robots \emph{$\mathcal{X}$} is denoted as 3, which are randomly located in the initial positions obtained by the LSTM-ARIMA model. The RIS is fixed at the center of the ceiling with a height of 3m, and the standard size of space is 8m and 6m. Additionally, there are four pillars with regular size 1m $\times$ 1m $\times$ 3m, two parterres with regular size 1m $\times$ 1m $\times$ 1m, and a fountain with a regular base size of 1.5m $\times$ 1.5m $\times$ 1m in the space. Note that, all the objects' heights mentioned above are more than that of the robot. The maximal transmit power at AP is pre-defined as 10 dBm, while the height of AP is defined as 2m. The number of reflecting elements in RIS $K$ and in sub-surface $\tilde{K}$ is defined as 20 and 5, while the total number of sub-surface $M$ is only increased linearly with $K$. The other simulation parameters are provided in Table.~\ref{Sim}. The performance of the proposed LSTM-ARIMA and D$^{3}$QN algorithms, the trajectories for all the robots, and the achievable sum-rate for all the robots are analyzed in the following sections.
\par
\vspace{-0.6cm}
\noindent
\begin{table}[!th]
  \caption{Simulation parameters \label{Sim}} 
  \centering
  \begin{tabular}{ccc}  
  \toprule 
  Parameter & Description & Value  \\ 
  \midrule
  C & Path loss when $d = 1$m & -30dB \\
  $\delta^{2}$ & Noise power variance& -75dBW \\ 
  $\overline{\alpha}_{Ai}$ & Path loss factor for AP-robot link& 3.5 \\ 
  $\overline{\alpha}_{Ri}$ & Path loss factor for RIS-robot link& 2.8 \\
  $\overline{\alpha}_{AI}$ & Path loss factor for AP-RIS link& 2.2 \\ 
  $\overline{T}$ & The replay memory capacity& 1000 \\ 
  $\mathbf{\mathcal{D}}$ & The number of episodes& 10000 \\ 
  $\overline{N}$ & The size of minibatch& 64 \\ 
  $\eta$ & Discount factor& 0.9 \\  
  $\psi$ & Learning rate& 0.05 \\   
  $\epsilon$ & Probability decision value & 0.1 \\      
  \bottomrule
  \end{tabular}  
\end{table}

\vspace{-0.3cm}
\subsection{Initial-final Position Pairs Predicted by LSTM-ARIMA}
\begin{figure}[htbp]
  \vspace{-0.5cm}
  \centering
  \begin{minipage}[t]{0.48\textwidth}
  \centering
  \setlength{\abovecaptionskip}{-0.05cm}
  \includegraphics[height=2.4in,width=3.2in]{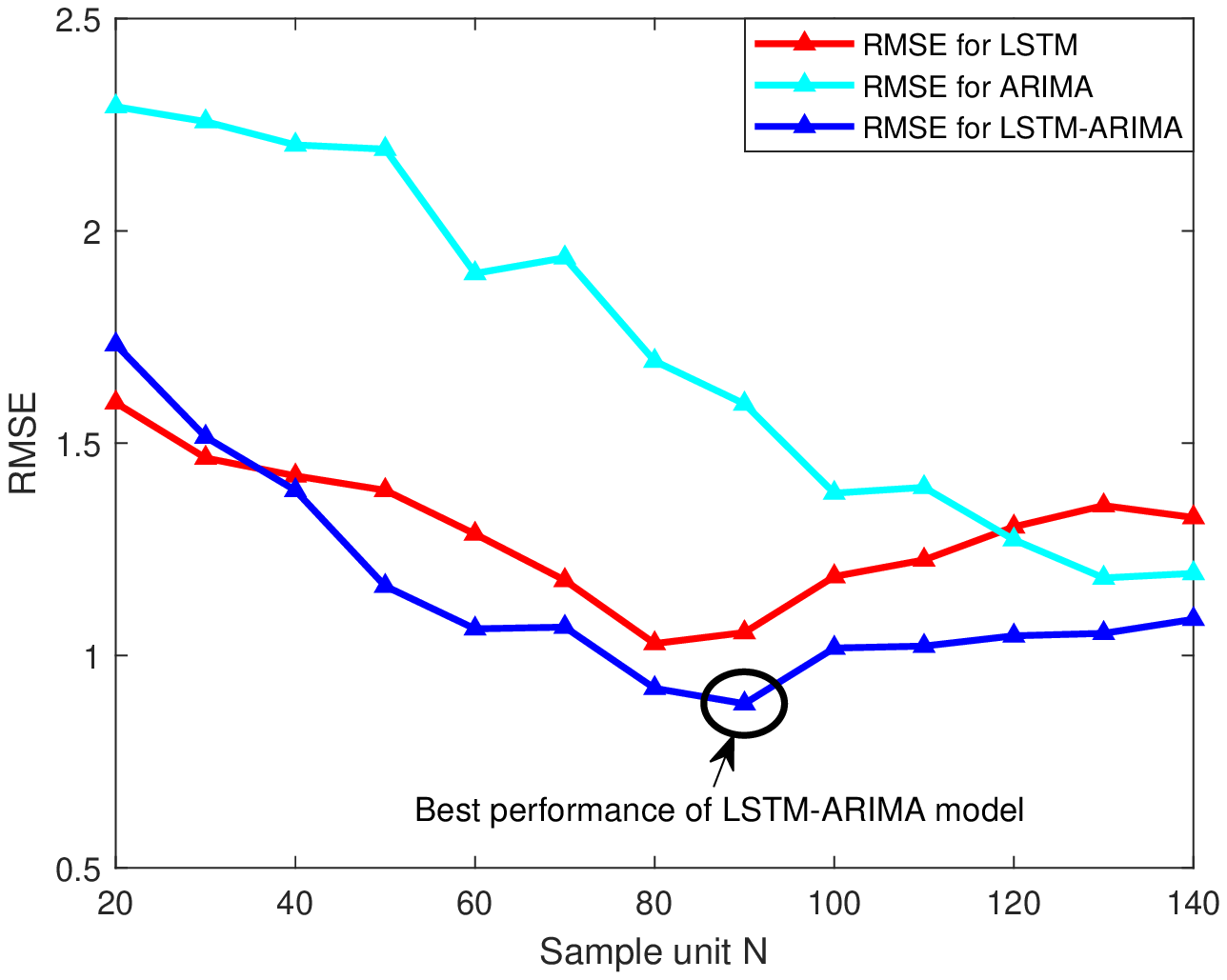}
  \caption{Prediction performance of different model.}
  \label{Performance_LSTM_ARIMA}
  \end{minipage}
  \begin{minipage}[t]{0.48\textwidth}
  \centering
  \setlength{\abovecaptionskip}{-0.05cm}
  \setlength{\belowcaptionskip}{-0.5cm}
  \includegraphics[height=2.4in,width=3.2in]{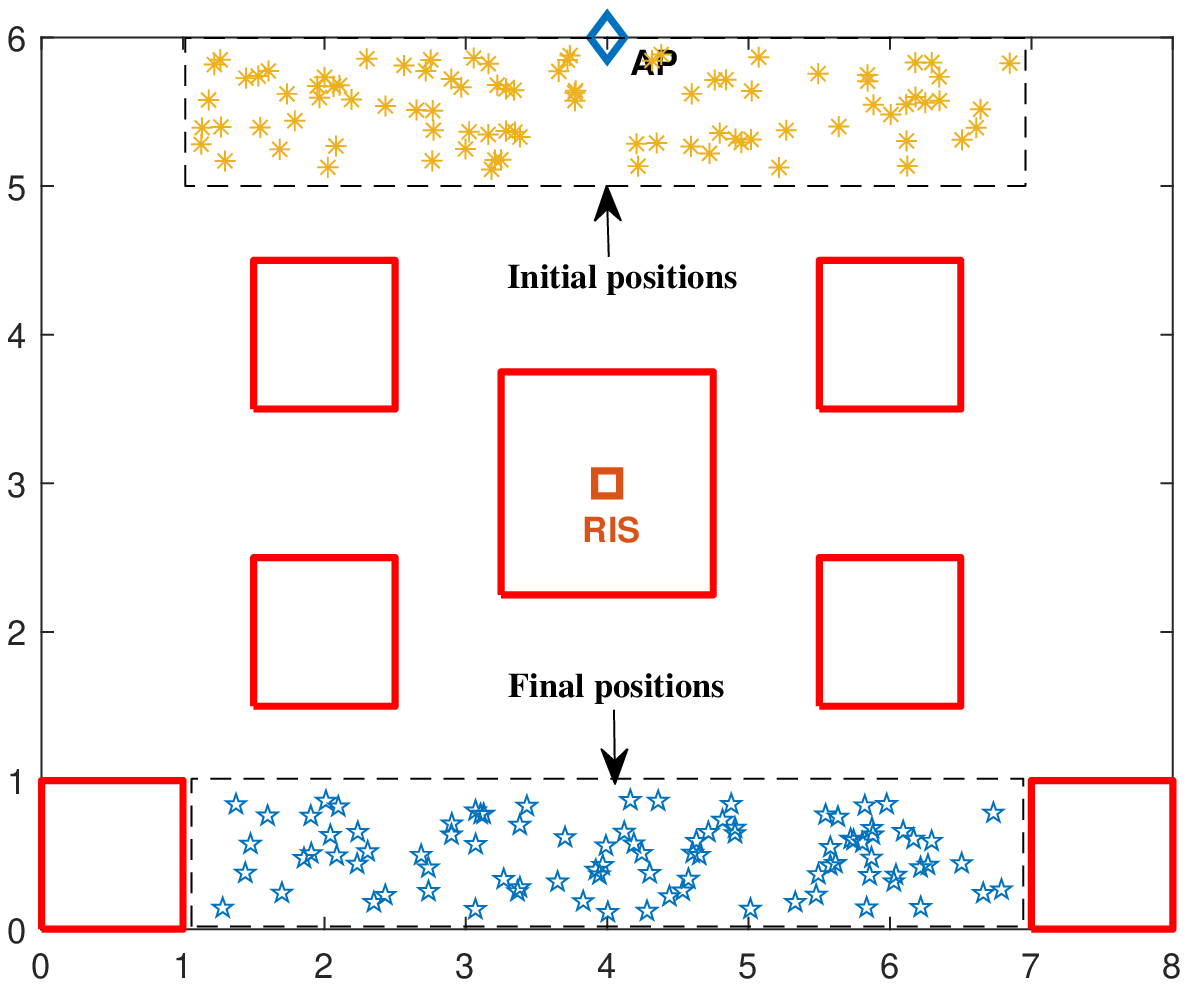}
  \caption{The range of optimal initial-final position pairs predicted by LSTM-ARIMA Model ($N=90$).}
  \label{LSTM_ARIMA1}
  \end{minipage}
\end{figure}

We adopted the LSTM-ARIMA model to predict the position pairs, where the position pairs are partitioned into initial positions and final positions groups. The possible initial position and final position are determined in the designated space, whose ranges are denoted as [1,7]m $\times$ [5,6)m and [1.1,6.9]m $\times$ (0,1]m. As shown in Fig.~\ref{Performance_LSTM_ARIMA}, the performance of the LSTM, ARIMA, and LSTM-ARIMA fusion models has been investigated under training samples by the different unit $N$ definitions. For LSTM, the changing trend of the RMSE proves the long sequence degrades its performance. It can be obtained that when the unit achieves $N=80$, LSTM is able to achieve the best performance than the other unit definition. For ARIMA, it is obtained that its performance keeps improving with $N$ increasing since it is able to handle long sequence issues. However, the historical data mentioned above is commonly regarded as a long unregular sequence, where LSTM and ARIMA have the capability of solving the unregular sequence and long sequence, respectively. Therefore, the dynamic weights \emph{$\overline{w}_{n}$} and \emph{$\tilde{w}_{n}$} are assigned to LSTM and ARIMA, which are dynamic determined for each predicted position pair. After fusion of the advantages of LSTM and ARIMA, when the unit is defined as $N=90$, LSTM-ARIMA achieves the best performance than other cases. As shown in Fig.~\ref{LSTM_ARIMA1}, the results for predicted initial and final position pairs are provided by the well-trained model ($N=90$). With the reliable model for initial-final position pairs prediction, the optimal trajectory with maximum sum-rate can be further explored.

\vspace{-0.5cm}
\subsection{Convergence for D$^{3}$QN Algorithm}
\begin{figure}[htbp]
  \centering
  \vspace{-0.5cm}
  \begin{minipage}[t]{0.48\textwidth}
    \centering
    \setlength{\abovecaptionskip}{-0.05cm}
    \includegraphics[height=2.4in,width=3.3in]{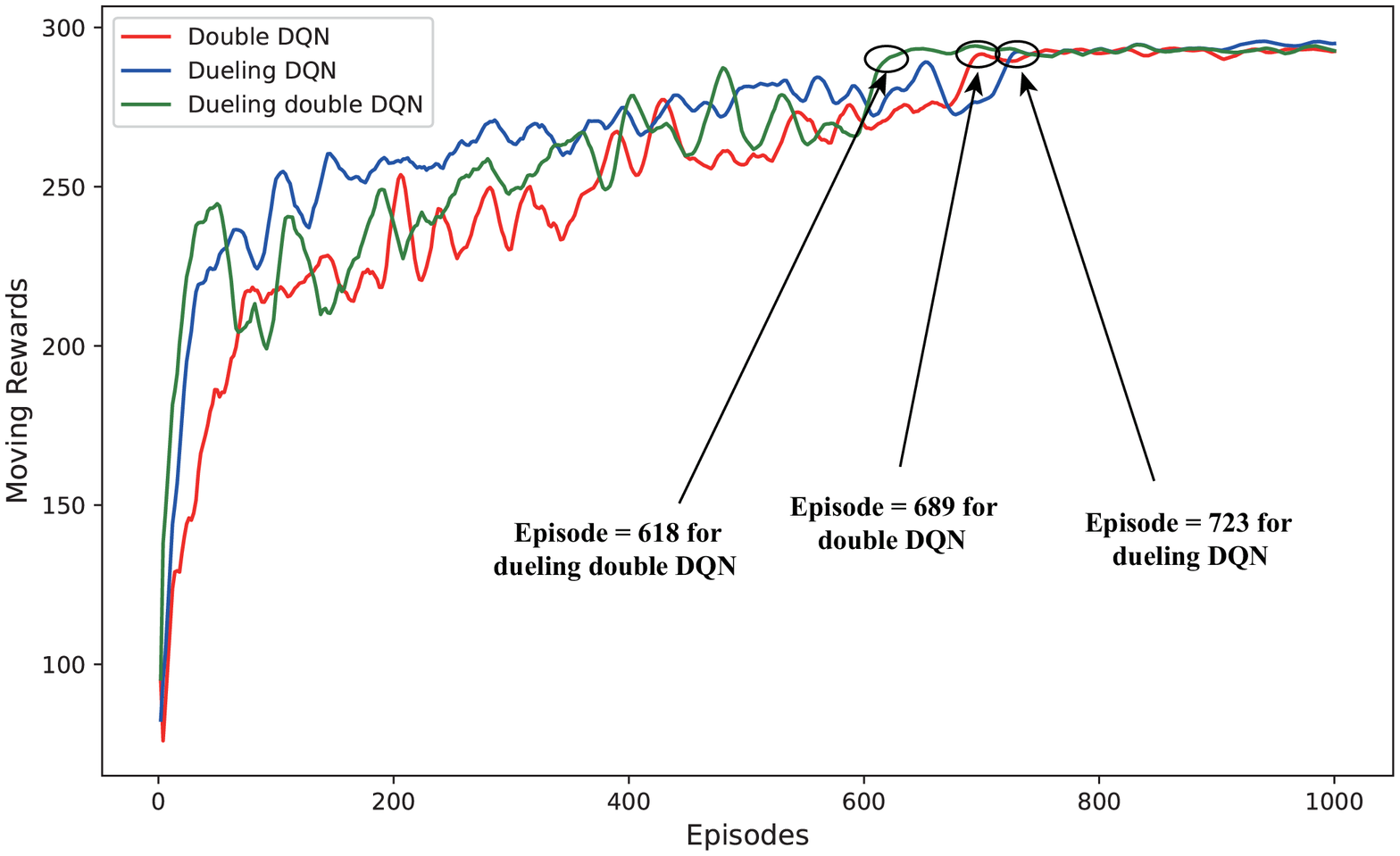}
    \caption{Performance for D$^{3}$QN algorithm comparing different algorithms.}
    \label{Performance for DDDQN algorithm comparing different algorithms}
  \end{minipage}
  \begin{minipage}[t]{0.48\textwidth}
  \centering
  \setlength{\abovecaptionskip}{-0.05cm}
  \setlength{\belowcaptionskip}{-0.5cm}
    \includegraphics[height=2.3in,width=3.2in]{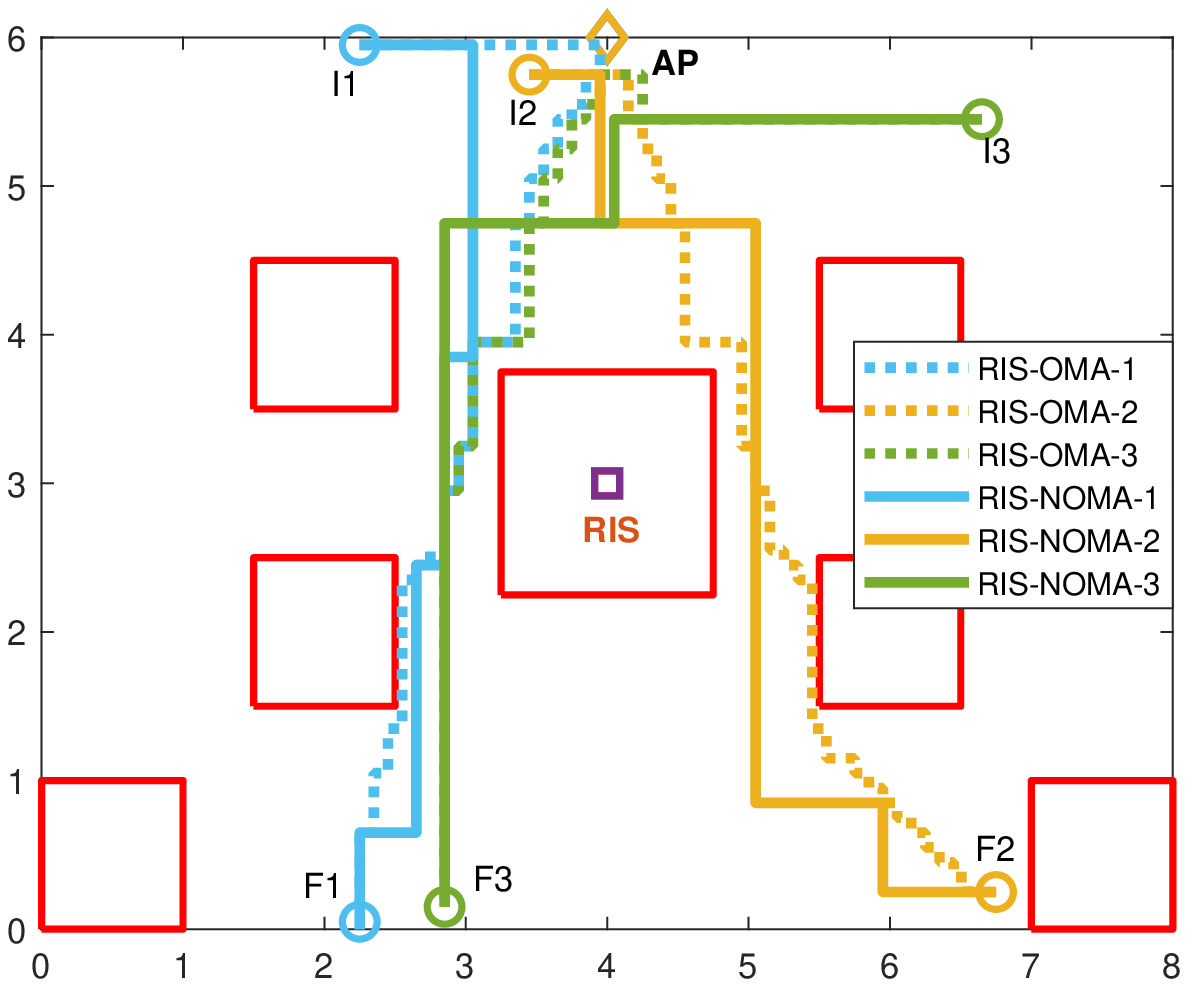}
    \caption{Trajectories for each robot under RIS-OMA and RIS-NOMA cases, $K$ = 30.}
    \label{path}
  \end{minipage}  
\end{figure}

The performance of the ML-based algorithm occupies a pivotal place in the entire optimization. For the proposed D$^{3}$QN algorithm, after selecting the initial-final positions of the robot, we optimize the sum-rate for all robots. In order to analyze the performance of D$^{3}$QN, we compared D$^{3}$QN algorithm to the double DQN algorithm and the dueling DQN algorithm with \emph{$\mathcal{X}=3$}, \emph{$\mathcal{P}=10dbm$}. As shown in Fig.~\ref{Performance for DDDQN algorithm comparing different algorithms}, the double DQN algorithm and dueling DQN algorithm have no big difference in convergence speed. They can converge when episodes are 689 and 723 respectively. However, the convergence speed of the proposed D$^{3}$QN is faster than these two algorithms, reaching 618. It is worth noting that in virtue of $\epsilon$-greedy strategy, the convergence episodes of these three algorithms cannot be guaranteed to be the same during each training. Therefore, the result given in the figure is the average convergence given by 10 repetitive training.

\vspace{-0.4cm}
\subsection{Trajectory and Achieved Sum-rate for Robots}
In this subsection, we provide the trajectory planning results for the robots. There are three pre-defined conditions for the robot's trajectory design. Firstly, denote the velocity of robots and the map resolution as 0m/s (stillness) or 0.1 m/s (movement), and 0.1 m, respectively, which guarantees each robot is able to move at most one cell at each timestep. Then, the size of a cell has been approximated to the center point of the cell which makes the robot arrive at the center of cells at each timestep and easily characterizes the received communication information. Finally, the movement direction of a robot is back, forth, stillness, left, and right. In the following, we discuss the results of trajectory design for the robots.

\subsubsection{Obtained Trajectories for Robots}
\begin{figure}[htbp]
  \vspace{-0.5cm}
  
  \centering
  \begin{minipage}[t]{0.48\textwidth}
  \centering
  \setlength{\abovecaptionskip}{-0.05cm}
  \includegraphics[height=2.4in,width=3.2in]{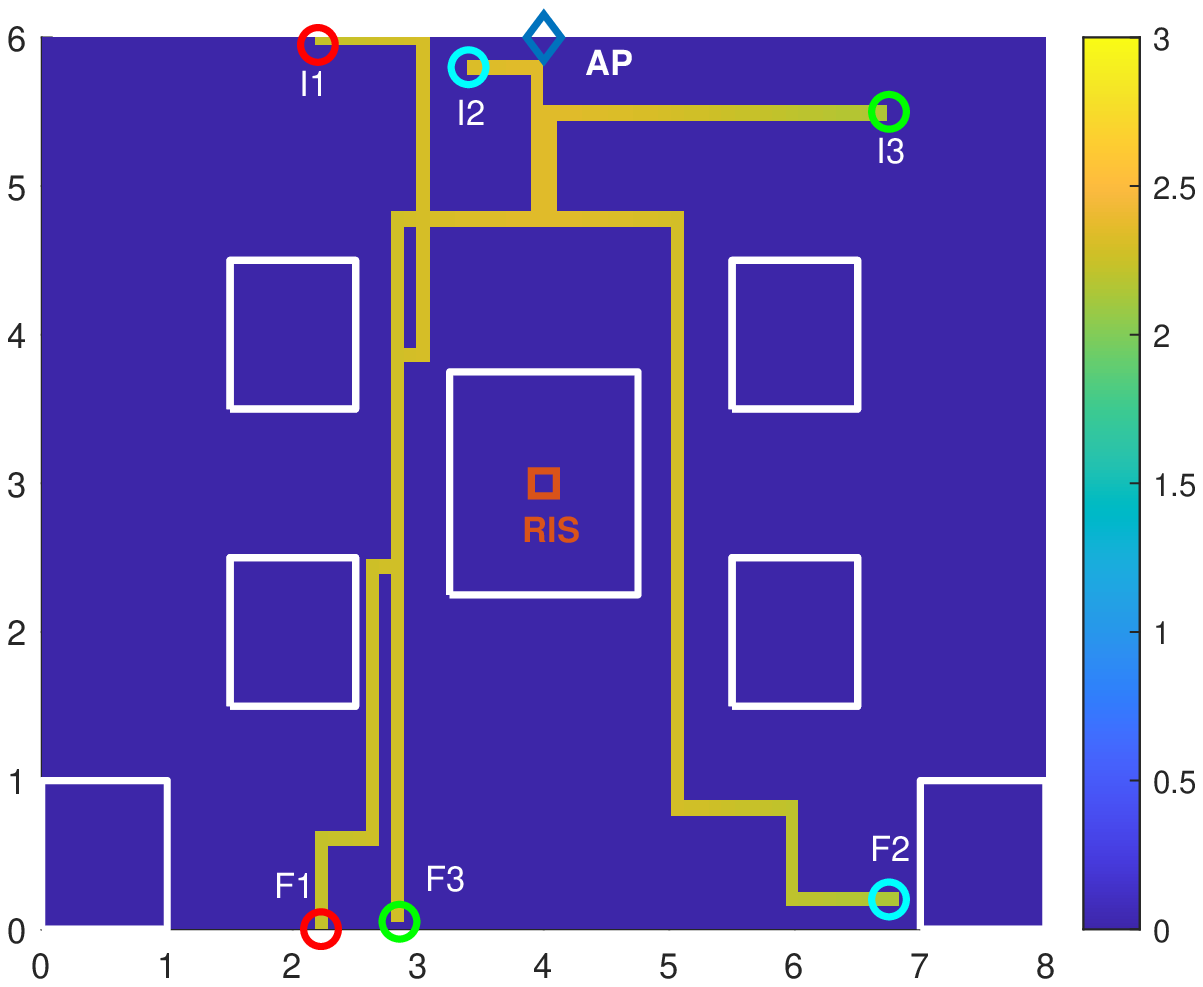}
  \caption{Communication rate for trajectories based on NOMA, $K$ = 30.}
  \label{RadioNOMA}
  \end{minipage}
  \begin{minipage}[t]{0.48\textwidth}
  \centering
  \setlength{\abovecaptionskip}{-0.05cm}
  \setlength{\belowcaptionskip}{-0.5cm}
  \includegraphics[height=2.4in,width=3.2in]{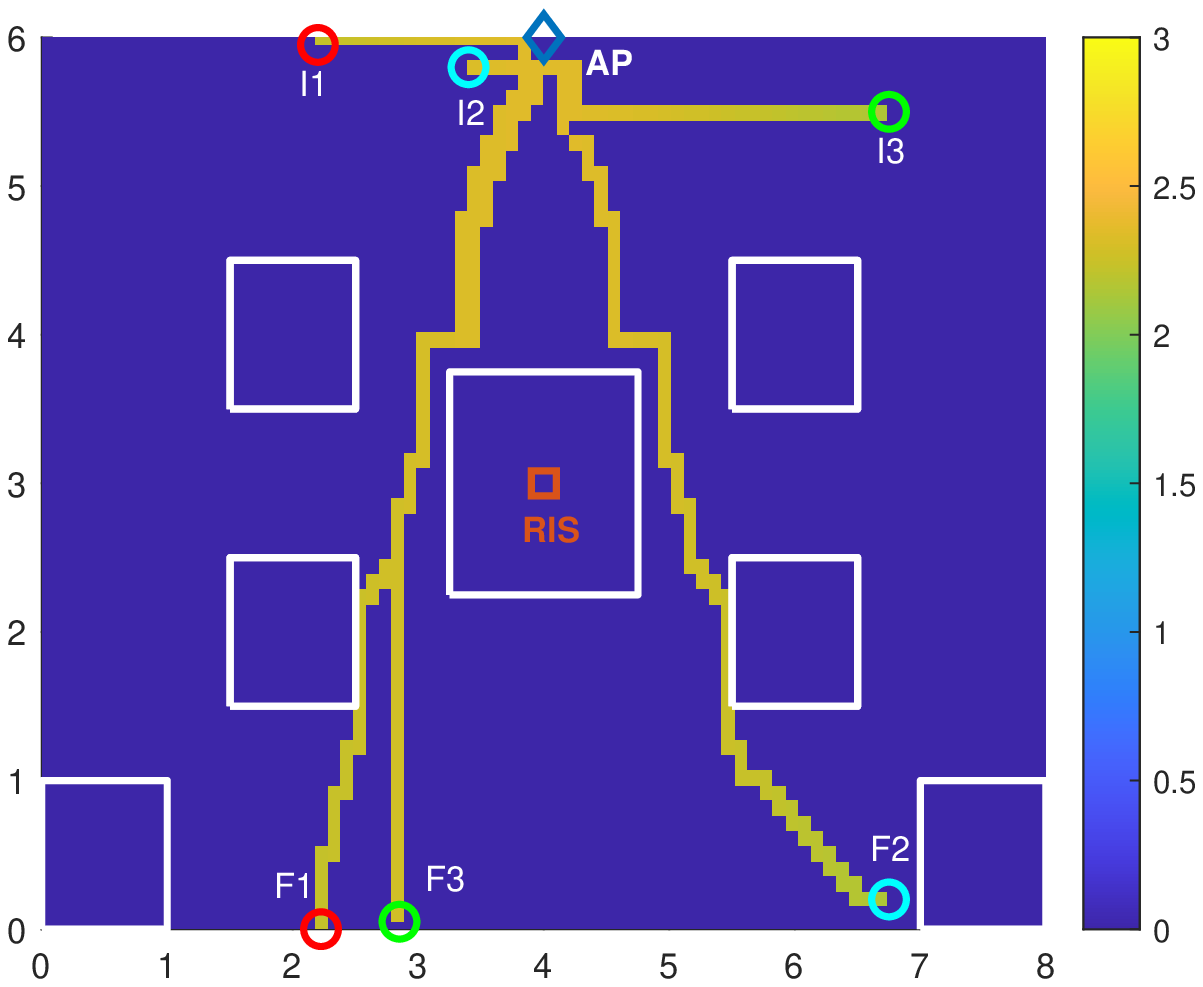}
  \caption{Communication rate for trajectories based on OMA, $K$ = 30.}
  \label{RadioOMA}
  \end{minipage}  
\end{figure}

As shown in Fig.~\ref{path}, the planned paths for all robots are depicted under "RIS-OMA" and "RIS-NOMA" cases, while the number of elements and sub-surfaces of the RIS is denoted as 30 and 6. The "$\circ$" with "$I_{\overline{w}},\overline{w}=\{1,2,3\}$" denotes the initial position for the robots, while the "$\circ$" with "$F_{\overline{w}},\overline{w}=\{1,2,3\}$" represents the final position. Moreover, dotted lines and solid lines with three colors are utilized to describe the paths under NOMA-aided and OMA-aided networks. The initial and final positions are the possible optimal pairs according to the proposed ML framework, which is randomly selected from the obtained range by LSTM-ARIMA and finally determined by the D$^3$QN algorithm. As shown in Fig.~\ref{RadioNOMA} and Fig.~\ref{RadioOMA}, the received communication rate is calculated on the robot located positions, which directly reflects the communication quality of the whole trajectories on each robot. In the two figures, the dark blue represents the area where robots are unselected or unreached, while the other colors indicate the received communication rate for robots' trajectories. Intuitively, it can be observed that the obtained trajectories by NOMA-aided and OMA-aided networks both tend to be close to RIS, while the shape of paths by NOMA-aided networks seems more regular than that of OMA-aided networks. One probable reason is, that for each robot, the OMA technology tends to achieve a reflected LOS-dominated channel model by traversing the cells covered by the RIS. The other reason is we aim to achieve the maximum long-term benefits in a time period instead of at each timestep. In other words, NOMA can improve the efficiency of trajectories design in a limited time instead of OMA.  Additionally, all trajectories are designed close to the RIS and AP,  which is able to verify the RIS is able to improve the received communication rate of robots.

\subsubsection{Impact of RIS on Robot Trajectory}
\begin{figure}[htbp]
  \vspace{-0.5cm}
  \centering
  \begin{minipage}[t]{0.48\textwidth}
  \centering
  \setlength{\abovecaptionskip}{-0.05cm}
  \includegraphics[height=2.4in,width=3.2in]{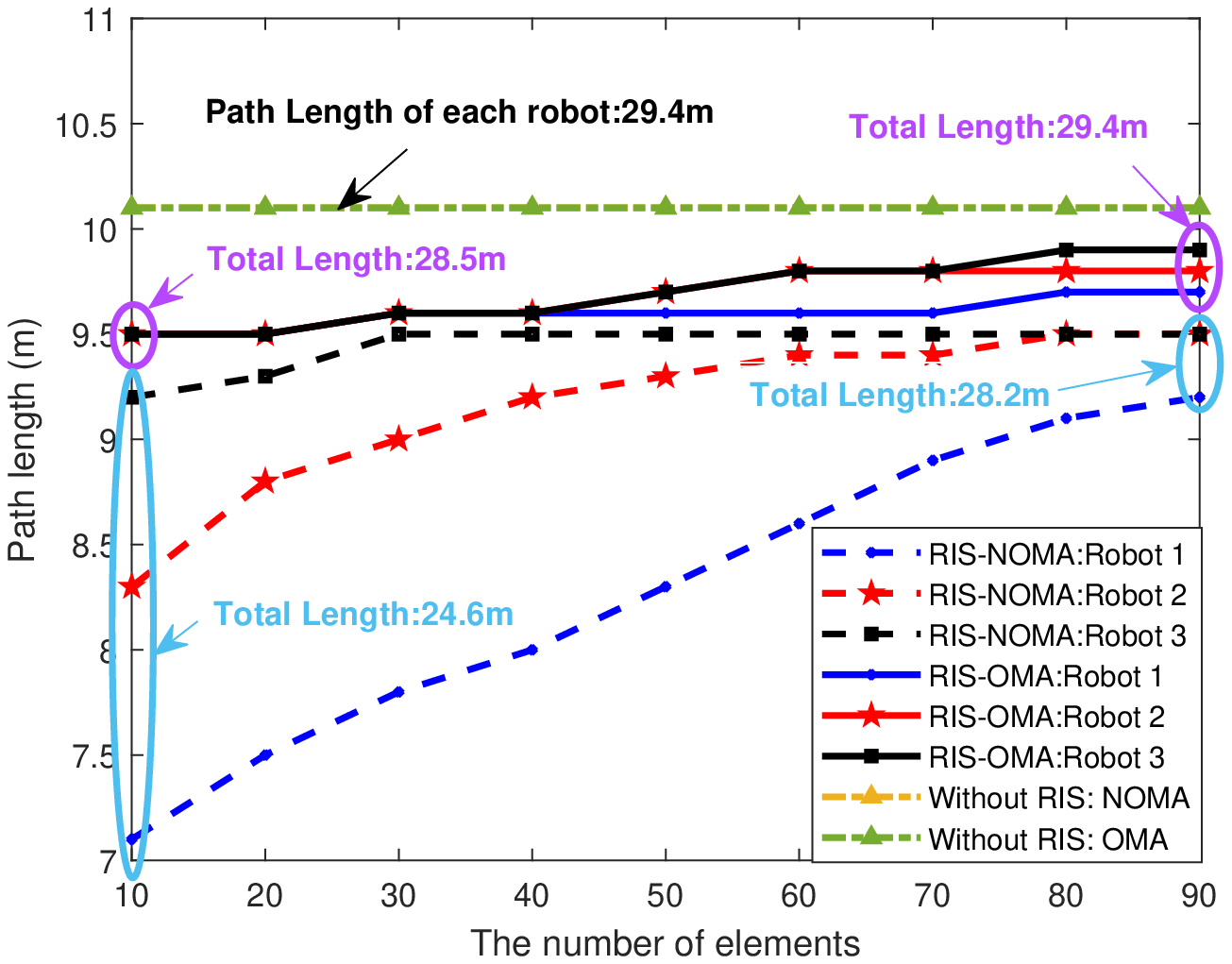}
  \caption{Path length for all robots with different number of elements in RIS.}
  \label{Pathlength}
  \end{minipage}
  \begin{minipage}[t]{0.48\textwidth}
    \centering
    \setlength{\abovecaptionskip}{-0.05cm}
    \setlength{\belowcaptionskip}{-1cm}
    \includegraphics[height=2.4in,width=3.2in]{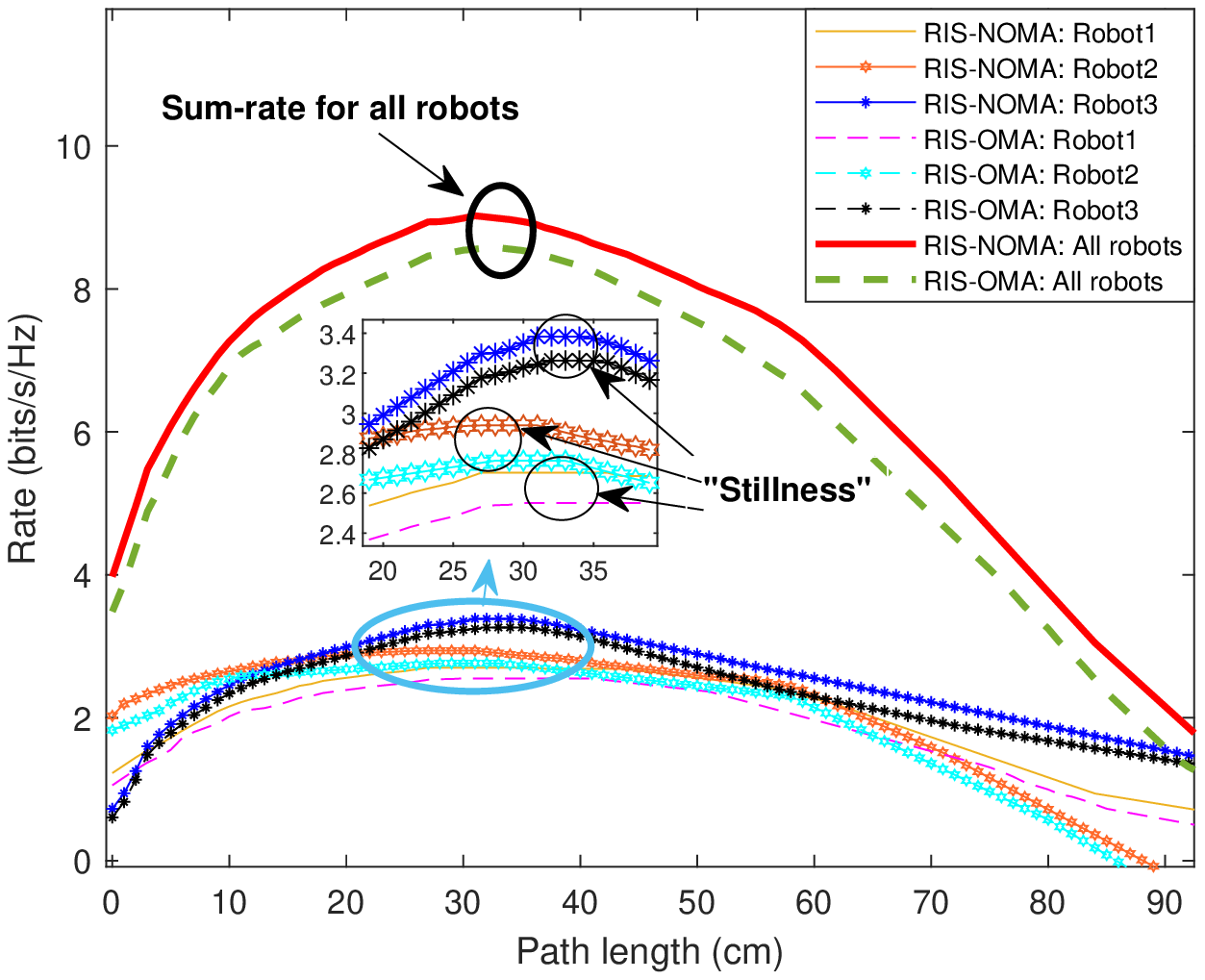}
    \caption{sum-rate versus path length with $K$=30.}
    \label{sum-rate}
  \end{minipage}  
\end{figure}

The employment of RIS brings an impact on trajectory design for the robots, which is demonstrated in Fig.~\ref{Pathlength} and Fig.~\ref{sum-rate}. In Fig.~\ref{Pathlength}, we compare ``RIS-NOMA" and ``RIS-OMA" cases,  where the designed total path length increased from 24.6m to 28.2m in NOMA networks, while it upgrades from 28.5m to 29.4m in OMA networks. This is because, with adding more elements in the RIS, the planned trajectories under the ``RIS-NOMA" case improve the overall communication quality of the environment, which makes robots have more spaces to explore. Additionally, it can be observed that the planned path length in the ``RIS-NOMA" case is longer than the path length in ``RIS-OMA" case, which indicates the ``stillness" are selected by robots at some states. Compared to the RIS deployment cases, the paths of robots in ``without RIS" cases are the longest, where the path length of each robot is the same in the NOMA and OMA networks. This is because the position of AP, positions of obstacles, simultaneously arriving conditions, and communication quality limitations make the length of the communication-sensitive path become different and much longer than the geographic path.
\par
In Fig.~\ref{sum-rate}, mark the "OMA" as a benchmark scheme, the maximal sum-rate for three robots at any positions on their designed trajectories is obtained. For the rate of robot 1 and robot 2, we can observe that they have the same rate at several adjacent positions. This is because they select ``stillness" at the current state, which proves they are constrained by the simultaneous arrival condition. These positions are the maximum rate position on the trajectory for each robot. ``stillness" guarantees to continue to receive high-quality communication information for robots while other robots still move. In the figure, it is observed that the maximum rate of position for each robot is 2.6m, 2.8m, 3.2m, robot 1 chooses ``stillness" 17 times while robot 2 chooses ``stillness" 4 times. It can be obtained that the total length of optimized trajectories for all robots is 7.8m, 9.1m, and 9.5m, while for all robots, the maximum sum-rate has been achieved when the total path length of each robot is 2.6m, 2.8m, and 3.1m. Finally, compared to the OMA-aided scheme, the ``RIS-NOMA" case outperforms the ``RIS-OMA" case.

\begin{figure}[htbp]
  \vspace{-0.5cm}
  \centering
  \begin{minipage}[t]{0.48\textwidth}
    \centering
    \setlength{\abovecaptionskip}{-0.05cm}
    \includegraphics[height=2.4in,width=3.2in]{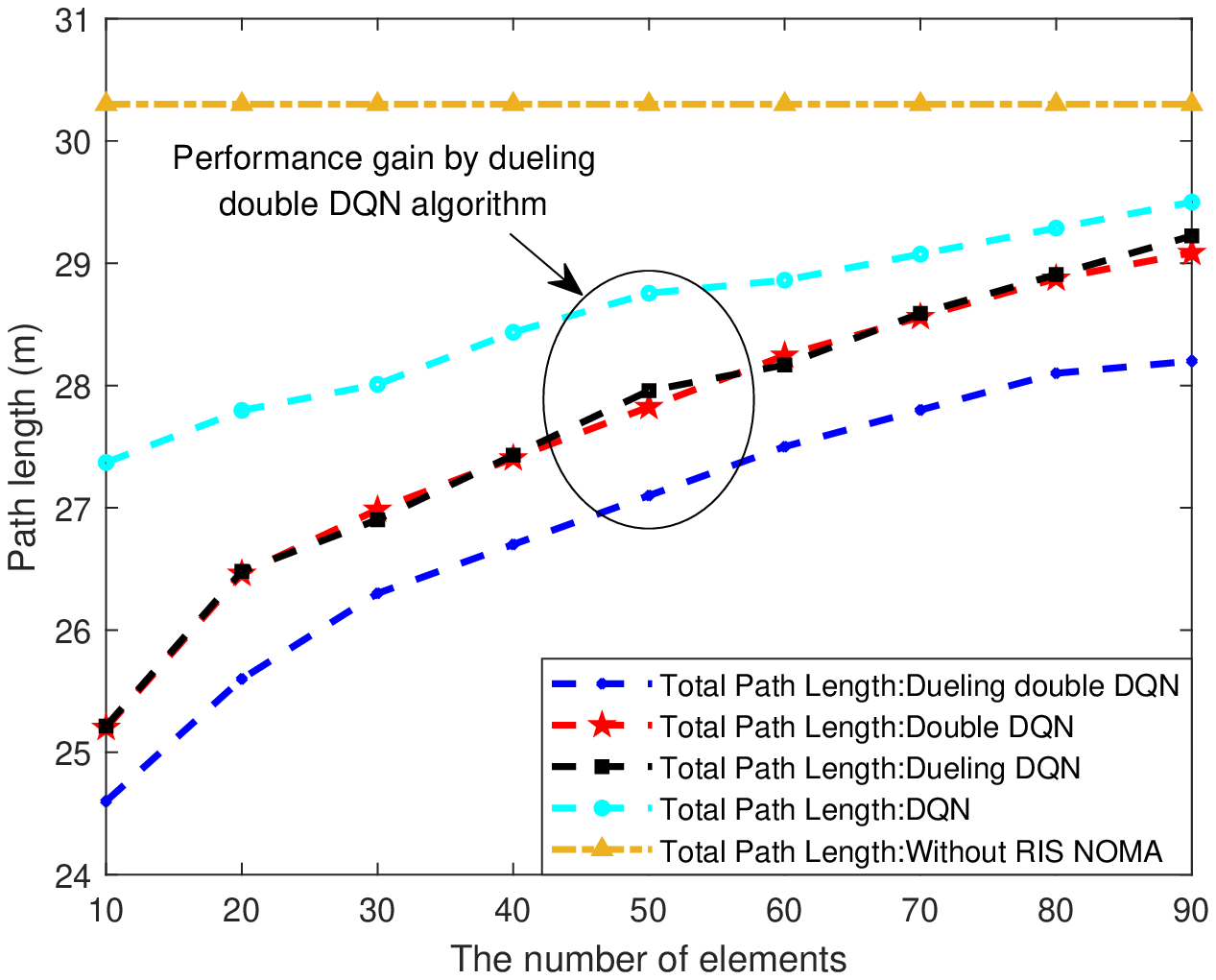}
    \caption{sum-rate versus path length with $K$=30 by different algorithms.}
    \label{length_al}
  \end{minipage}
  \begin{minipage}[t]{0.48\textwidth}
    \centering
    \setlength{\abovecaptionskip}{-0.05cm}
    \setlength{\belowcaptionskip}{-0.4cm}
    \includegraphics[height=2.4in,width=3.2in]{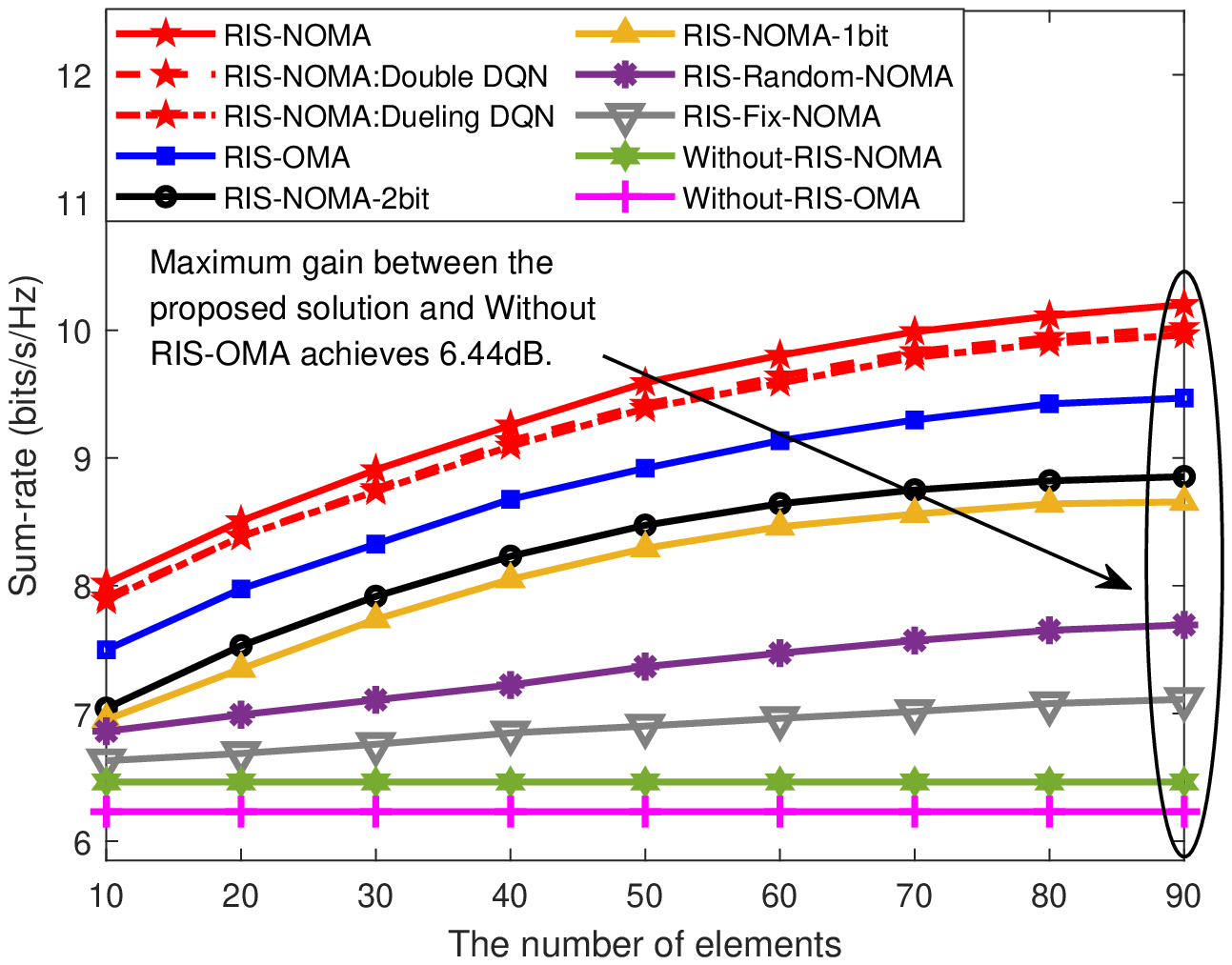}
    \caption{Sum-rate of whole trajectories for all robots under different number of elements.}
    \label{Elements}
  \end{minipage}  
\end{figure}

In Fig.~\ref{length_al}, the comparison of the proposed algorithm, double DQN algorithm, dueling DQN algorithm, and DQN is presented, where the different elements in the RIS will influence the obtained total path length. It can be observed that the proposed D$^3$QN algorithm is able to find a shorter total path than the conventional ML algorithms. This is because the structure of the proposed D$^3$QN algorithm significantly supplements the limitations of the conventional ML algorithm. Therefore, it can be obtained that the performance of the D$^3$QN algorithm outperforms the conventional ML algorithms. Additionally, the ML solutions for the RIS-NOMA case all achieve a shorter total path length than the ``Without RIS: NOMA" case. The effectiveness of RIS on total path length optimization has been proved.

In Fig.~\ref{Elements}, the sum-rate of all robots' trajectories is demonstrated, where the different elements in the RIS will influence the obtained communication sum-rate of robots. In order to verify the benefits brought by the RIS, we compare ``RIS-NOMA" with benchmarks of ``RIS-OMA", ``RIS-NOMA-2bit", ``RIS-NOMA-1bit", ``RIS-Random-NOMA", ``RIS-Fix-NOMA", ``Without-RIS-NOMA", and ``Without-RIS-OMA" cases. It can be observed that the sum-rate performances of all considered RIS-aided schemes increase with the increase of elements. The performances of ``RIS-NOMA" significantly outperform the other seven benchmark schemes. The achieved sum-rate by adjusting the RIS phase shift with higher resolution bits outperforms the lower resolution bits, while the gap between the ``RIS-NOMA" and the RIS-aided low phase shift resolution case becomes larger when the elements increase. It is observed that the performance achieved by optimizing RIS phase shifts outperforms the ``RIS-Random" case whose reflection coefficients are randomly set. For ``RIS-Fix-NOMA", the phase shift of RIS is fixed and the long-term power allocation is optimized, it can be obtained that a significant performance degradation occurred compared to jointly optimizing the phase shift and power allocation. Finally, it can be observed that all RIS-NOMA cases outperform the RIS-OMA cases, while the maximum gain between the proposed solution and Without RIS-OMA achieves 6.44dB. Additionally, the performance of different algorithms on the sum-rate optimization is also considered, where the proposed algorithm outperforms the double DQN and dueling DQN algorithms. The performance of the double DQN and dueling DQN algorithms is close. Considering the results for path length and the maximum sum-rate of the path, it can be obtained that employing NOMA technology is able to strike a balance between the maximum sum-rate exploration and the shortest path design between the initial position and final position.
\par
\begin{figure}[htbp]
  \setlength{\belowcaptionskip}{-0.5cm}
  \centering
  \includegraphics[height=2.4in,width=3.2in]{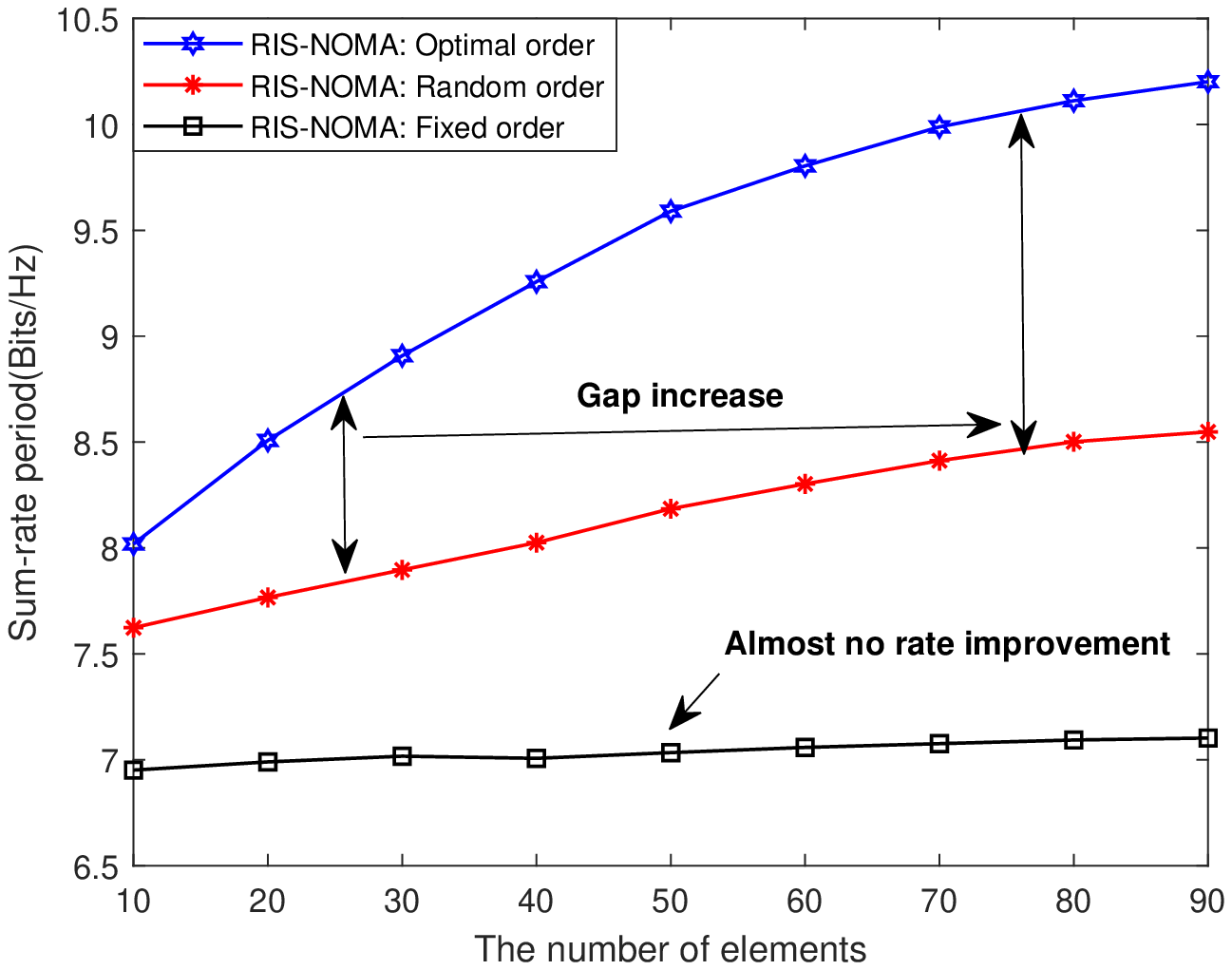}
  \caption{Sum-rate versus decoding order under different elements numbers of RIS.}
  \label{Trajectory_order}
\end{figure}

In Fig.~\ref{Trajectory_order}, to explore the influence brought by decoding order, we compare three schemes based on the different numbers of clusters: optimal order, random order, and fixed order. Optimal order denotes the best order selected by the D$^{3}$QN-based algorithm. As shown in the Figure, the DQN-based optimal order scheme significantly outperforms the random order scheme and fixed order, which highlights the necessity of exploring the optimal decoding order. Additionally, it can be observed that the gaps among the optimal schemes and the random order schemes present a huge difference with the increase of elements. The fixed order becomes insensitive to sum-rate improvement, which keeps very slightly changing with the elements increase. This is because optimal order schemes provide the optimal power allocated policy for each robot, which is ignored in random order schemes and fixed schemes. 

\vspace{-0.2cm}
\begin{remark}\label{remark 4}
  The optimal decoding order is found by an exhaustive search algorithm, which costs at least $\mathcal{X}!$ iterations, which is better to be invoked in the scenario with a small number of decoding users, otherwise, it will bring high complexity to the whole algorithm.
\end{remark}

\vspace{-0.5cm}
\section{Conclusion}
\vspace{-0.1cm}
In this paper, we explored a downlink RIS-aided multi-robot NOMA network. The robot trajectories' sum-rate maximization problem was formulated by jointly optimizing trajectories for robots, reflecting the coefficients matrix, the decoding order, and the power allocation at the AP, subject to the QoS for all the robots. To tackle the formulated problem, a novel machine learning scheme was proposed, which combines the LSTM-ARIMA model and D$^{3}$QN algorithm. LSTM-ARIMA algorithm is employed to predict the possible initial and final positions for robots, while the D$^{3}$QN algorithm is invoked to plan optimal trajectories for the robots and design the phase shift matrix, determining the optimal initial and final positions for robots. Numerical results were provided for demonstrating that the proposed RIS-aided NOMA networks achieve significant gains compared to RIS-OMA and without-RIS-aided schemes. The investigated LSTM-ARIMA and D$^{3}$QN algorithms attained considerable performance compared to the vanilla ML algorithm. In a real application, to improve the efficiency of the communication system, energy is another key element that needs to be considered. Therefore, striking a tradeoff between energy and sum-rate is a potential research topic in future work \cite{Zan}. Moreover, with respect to the multi-robot system in the indoor environment, deploying a single RIS may not be efficient to establish LOS links due to the dense obstacles. In this case, the multiple distributed RISs can be considered to leverage the richer reflected paths to further improve performance \cite{Mei1, Mei2}.

\appendices

\ifCLASSOPTIONcaptionsoff
  \newpage
\fi

\end{document}